\documentclass{article} % For LaTeX2e
\usepackage{iclr2025_conference,times}
\usepackage{multirow}
\usepackage{multicol}
\usepackage{amsmath,amsfonts,bm}

\def\eqref#1{equation~\ref{#1}}
\def\1{\bm{1}}

\def\vc{{\bm{c}}}

\def\vg{{\bm{g}}}

\def\vk{{\bm{k}}}

\def\vp{{\bm{p}}}
\def\vq{{\bm{q}}}

\DeclareMathAlphabet{\mathsfit}{\encodingdefault}{\sfdefault}{m}{sl}
\SetMathAlphabet{\mathsfit}{bold}{\encodingdefault}{\sfdefault}{bx}{n}

\usepackage{booktabs}
\usepackage{hyperref}
\usepackage{makecell}
\usepackage{wrapfig}  
\usepackage{adjustbox}
\usepackage{mathrsfs}
\usepackage{listings}
\usepackage{pifont}
\usepackage{graphicx}
\usepackage{subfigure}
\usepackage{alltt}  
\usepackage{subcaption}
\usepackage{url}
\usepackage{pifont}
\usepackage{ulem}
\usepackage{svg}
\usepackage{makecell}
\usepackage{cleveref}
\crefname{section}{§}{§§}
\Crefname{section}{§}{§§}
\usepackage{svg}
\usepackage{colortbl}
\usepackage{fontawesome}

\usepackage[figuresright]{rotating}
\usepackage[most]{tcolorbox}
\definecolor{BoxBackground}{RGB}{240, 240, 240} % 浅灰色背景
\definecolor{BoxFrame}{RGB}{0, 0, 0} % 黑色边框
\definecolor{TitleBackground}{RGB}{0, 0, 0} % 标题背景颜色
\definecolor{TitleText}{RGB}{255, 255, 255} % 标题文字颜色
\tcbset{
  academicbox/.style={
    boxsep=5pt,
    left=2pt,
    right=2pt,
    bottom=0.5pt,
    boxrule=0.5pt,
    colback=BoxBackground,
    colframe=BoxFrame,
    colbacktitle=TitleBackground,
    coltitle=TitleText,
    enhanced,
    attach boxed title to top left={yshift=-0.1in,xshift=0.1in},
    boxed title style={boxrule=0pt,colframe=white},
    title={#1},
  }
}
\newtcolorbox{AcademicBox}[1][]{academicbox=#1}

\title{LOGO --- Long cOntext aliGnment via efficient preference Optimization}

\author{Zecheng Tang,\quad Zechen Sun,\quad Juntao Li\thanks{Corresponding Author},\quad Qiaoming Zhu,\quad Min Zhang\\
School of Computer Science and Technology, Soochow University \\
\texttt{\{zctang,zcsuns\}@stu.suda.edu.cn},\quad \texttt{\{ljt,qmzhu,minzhang\}@suda.edu.cn}\\
}

\iclrfinalcopy

\begin{document}

\maketitle

\vspace{-2em}

\begin{center}
    \textbf{\textit{\faGithub~Code \& Data: \textcolor{violet}{ \url{https://github.com/ZetangForward/LCM_Stack.git}}}}
\end{center}

\begin{abstract}
Long-context models~(LCMs) have shown great potential in processing long input sequences~(even more than 100M tokens) conveniently and effectively.
With significant progress, recent research has pointed out that LCMs can accurately locate token-level salient information within the context.
Yet, the generation performance of these LCMs is far from satisfactory and might result in misaligned responses, such as hallucinations.
To enhance the generation capability of LCMs, existing works have investigated the effects of data size and quality for both pre-training and instruction tuning.
Though achieving meaningful improvement, previous methods fall short in either effectiveness or efficiency.
In this paper, we introduce LOGO~(Long cOntext aliGnment via efficient preference Optimization), a training strategy that first introduces preference optimization for long-context alignment.
To overcome the GPU memory-bound issue caused by the long sequence, LOGO employs a reference-free preference optimization strategy and adopts a position synthesis method to construct the training data.
By training with only 0.3B data on a single 8$\times$A800 GPU machine for 16 hours, LOGO allows the Llama-3-8B-Instruct-80K model to achieve comparable performance with GPT-4 in real-world long-context tasks while preserving the model's original capabilities on other tasks, e.g., language modeling and MMLU.
Moreover, LOGO can extend the model's context window size while enhancing its generation performance.
\end{abstract}

% \begin{figure}[ht]
%     \centering
%     \vspace{-0.4cm}
%     \subfigure[Model performance on long-context tasks]{
%         \includegraphics[width=0.49\linewidth]{figure/intro2_0.pdf}
%         \label{fig:i0}
%     }
%     % \hspace{0.01\linewidth}
%     \subfigure[Trends of retrieval score and recall score]{
%         \includegraphics[width=0.46\linewidth]{figure/intro2.pdf}
%         \label{fig:i1}
%     }
%     \vspace{-0.2cm}
%     \caption{(a) The overall performance of Long-context Models on real-world long-context tasks. (b) Trends of retrieval score and recall score of Llama-3-8B-Instruct-80K model in the training phase.}
%     \label{fig:intro}
%     \vspace{-0.3cm}
% \end{figure}

\begin{figure}[ht]
    \centering
    \vspace{-1em}
    \includegraphics[width=\linewidth]{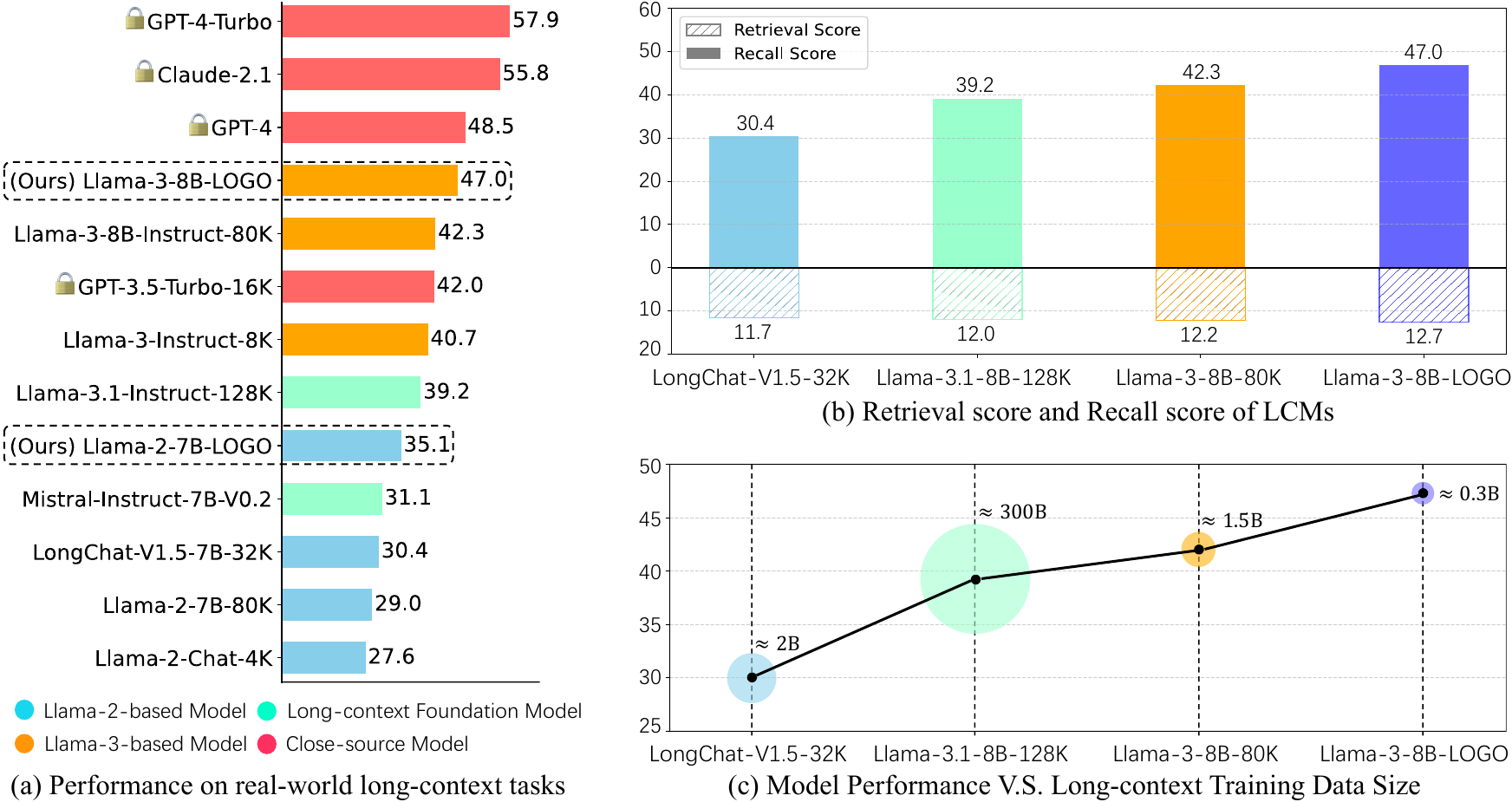}
    \caption{(a) Performance of LCMs on real-world long-context tasks; (b) Retrieval score~(long-context understanding ability) and recall score~(generation ability) of LCMs on the synthetic retrieval long-context task~(multi-value NIAH); (c) Long-context (pre-)training data size for each LCM.}
    \label{fig:intro}
    \vspace{-1.5em}
\end{figure}

\section{Introduction}
\label{sec:intro}
With the rapid advancements of Large Language Models~(LLMs), handling long contexts~(even more than 100M tokens~\citep{claude_3_5}) has become a fundamental capability for recent LLMs.
This further unlocks the potential of LLMs for novel tasks and applications, e.g., code analysis~\citep{zhu2024deepseek}, while simultaneously eliminating the need for complex toolchains and intricate workflows that were previously required to overcome the context-length constraints~\citep{ravaut2024context}.

Yet, recent studies have pointed out that these long-context models~(LCMs) failed to achieve satisfactory performance in long-context tasks, where LCMs might produce misaligned results, such as instruction unfollowing and hallucinations~\citep{belyi2024luna,zhang2024longcite}.
To mitigate the above issue, the open-source community has made significant efforts, primarily focusing on building high-quality long instruction data and extending the data size~\citep{wu2024long, bai2024longalign,fu2024data,bai2024longalign}.
As shown in Fig.~\ref{fig:intro}, though achieving meaningful improvement, these methods fall short in effectiveness or efficiency. 
For instance, the Llama-3.1-8B-128K model~\cite{llama3_1modelcard} was pre-trained on around 300B long instruction data, but it even underperforms the Llama-3-8B-Instruct-80K model~\citep{zhang2024extending}, which was post-trained with 1.5B high-quality long instruction data based on the Llama-3-8B-Instruct model~\citep{llama3modelcard}.
As for the Llama-3-8B-Instruct-80K model, it shows slight improvement compared to the baseline and still lags greatly behind the closed-source counterparts like GPT-4~\citep{achiam2023gpt}.

Recently, \citet{wu2024retrieval} pointed out that LCMs can accurately locate token-level salient information within the context.
As shown in Fig.~\ref{fig:intro}(b), we visualize the information retrieval capability\footnote{Retrieval capability is reflected through the recall score of salient tokens located by retrieval heads~\citep{wu2024retrieval}. We calculate the average recall score across the top-10 retrieval heads. A higher retrieval score indicates that the LCM can retrieve more critical information. Details are shown in Appendix~\ref{appdix:into}.}~(reflected by the retrieval score) and the generation capability~(reflected by the recall score) of different LCMs on the synthetic retrieval task, where we can observe a minimal difference among the retrieval scores from various LCMs, but large differences in their generation performance.
This suggests that while LCMs are adept at identifying key information within long contexts, they struggle to effectively utilize the retrieval information for generation.
The underlying cause might be the commonly used training approach of LCMs, which relies on token-level maximum likelihood loss, i.e., Cross-Entropy~(CE) loss, calculated on both the context and the predictions. 
Given that the context's sequence length is typically much longer than the prediction portion, the feedback signal~(CE loss) from the prediction is often overshadowed by that from the context.
As a result, the CE loss becomes ineffective in optimizing the generation capabilities of LCMs.

% During training, CE loss is applied to both the context and the predictions. 
% However, the sequence length of context is usually much longer than the prediction part, as a result of which the loss feedback of the prediction part is overwhelmed by that of the context.
% Thus, the CE loss is ineffective in optimizing the generation capability of LCMs.
% While this approach aids in better language modeling over long sequences, the loss attributed to the long context tends to dominate the overall loss distribution. 
% As a result, the minimal loss assigned to the prediction fails to adequately guide the model towards correct outputs. 
% Furthermore, CE loss can only teach the model what is correct, rather than preventing models from generating erroneous (misaligned) outputs. 
% training strategy that aims to teach the LCMs to distinguish between preference predictions~(i.e., correct outputs) and dis-preference predictions (i.e., misaligned outputs like hallucinations).

To effectively optimize LCMs for generating desired outputs and avoid misaligned results, this paper introduces \textbf{LOGO}~(\textbf{L}ong c\textbf{O}ntext ali\textbf{G}nment via efficient preference \textbf{O}ptimization), the first training strategy that incorporates preference optimization for long-context alignment.
There are two key components in LOGO: (1) a training objective designed to guide LCMs to distinguish between preference predictions~(i.e., correct outputs) and dis-preference predictions~(e.g., misaligned outputs like hallucinations), and (2) a corresponding data construction pipeline that only involves open-source models.
It is worth noting that training with long sequence data is a memory-intensive task~\citep{dao2023flashattention} and the DPO algorithm also has a high GPU memory demand.
To overcome the GPU memory-bound and improve the training efficiency, LOGO adopts a reference-free training objective and the positional indices synthesis method~\citep{zhu2023pose}.
Consequently, we can perform the LOGO training with only 0.3B data on a single 8$\times$A800 GPU machine within 16 hours.

By training with LOGO, LCMs can achieve significant improvements in real-world tasks and gain moderate improvements in synthetic and language modeling tasks, as well as maintaining good performance on the short-context tasks, e.g., MMLU~\citep{hendrycks2020measuring}.
As shown in Figure~\ref{fig:intro}(a), our Llama-3-8B-LOGO significantly outperforms GPT3.5-Turbo in real-world tasks and approaches the performance of some top closed-source models like GPT-4.
Additionally, LOGO can also generalize to the training of short-context LLMs such as Llama-2-7B-Chat-4K~\citep{touvron2023llama}, which can potentially extend their context window size up to 8 times~(e.g.,32K context window size for Llama-2-7B-Chat-4K) while simultaneously enhancing their performance substantially.

\section{Related Work}
\label{sec:related_work}
\vspace{-0.5em}
\subsection{Long Context Scaling and Long Context Alignment}
Two steps are essential for empowering LLMs with the ability to handle long-context tasks: 1) context scaling, which expands the limited context window size to support long-context tasks, e.g., from 8k to 128k; and 2) long-context alignment, which ensures that LCMs can follow long instructions. 
Currently, the open-source community mainly focuses on the former, primarily by (1) post-training models on long instruction data~\citep{chen2023longlora,xiong2023effective,fu2024data,zhang2024extending}, (2) devising novel model architectures~\citep{yang2023gated,zhang2024sinklora,tworkowski2024focused}, and (3) modifying positional encoding~\citep{peng2023yarn,chen2023extending,jin2024llm} to extend the context window of LLMs.
However, current works~\citep{belyi2024luna,hsieh2024ruler,zhang2024longcite} indicated that LCMs still underperform in long-context tasks, frequently manifesting issues such as hallucinations and failure to follow instructions, despite possessing large context window size.
To mitigate this issue, \citet{bai2024longalign} and \citet{wu2024long} proposed to align the LCMs in long-context scenarios by synthesizing long-dependency instruction data to fine-tune the models. 
Some LLMs are even pre-trained with massive long instruction data~\citep{jiang2023mistral,dubey2024llama,abdin2024phi}.
Yet, despite numerous attempts that have been made to improve the data quality and quantity, the performance of open-source LCMs still lies far behind close-source LCMs.
Therefore, focusing solely on data augmentation methods can not resolve the long-context alignment problem efficiently and effectively.
In this work, we address the above issue from the training objective perspective. 
Building upon the language modeling task, we introduce LOGO, which contains a long-context preference optimization training objective. 
Experimental results demonstrate that, with a small amount of data and computational resources, LOGO can significantly enhance the generation capability of LCMs.

\subsection{Model Alignment with Direct Preference Optimization}
Direct Preference Optimization~(DPO)~\citep{rafailov2024direct} is a widely adopted RLHF algorithm~\citep{ouyang2022training} that aims to align models with human preferences.
Compared to other reinforcement learning methods, e.g., PPO~\citep{schulman2017proximal}, DPO can achieve strong performance while eliminating the need for a separate reward model. 
Unlike Supervised Fine-Tuning~(SFT), which guides LLMs to fit predictions to ground truth at the token level, DPO updates the model parameters with discrete evaluation scores.  
Specifically, DPO teaches the model to ``reject'' misaligned responses and ``accept'' preferred responses with differently assigned prediction scores. 
Significant efforts have been made to enhance the effectiveness and efficiency of DPO, such as CPO~\citep{xu2024contrastive}, TPO~\citep{saeidi2024triple}, and ORPO~\citep{hong2024reference}. 
Among them, SimPO~\citep{meng2024simpo} utilizes the average log probability of a sequence as the implicit reward, which better aligns with the generation tasks and eliminates the need for a reference model. 
\section{Methodology}
\label{sec:methodology}

\subsection{Background}
\label{subsec:background}

\paragraph{Direct Preference Optimization~(DPO) and Simple Preference Optimization~(SimPO)}
DPO is one of the most popular offline preference optimization strategies in RLHF~\citep{rafailov2024direct}.
Given prompt $x$, DPO aims to maximize the likelihood of a preferred response $y_w$ over a dis-preferred one $y_l$, thereby preventing the model from generating undesired content.
There are three essential modules in the DPO training process: one reference model and one policy model for calculating the DPO loss jointly, and one evaluation strategy~(or evaluation model) for distinguishing between $y_{w}$ and $y_{l}$.
SimPO~\citep{meng2024simpo} is an improved variant of DPO, which employs an implicit reward formulation that directly aligns with the generation metric, e.g., PPL, thereby eliminating the need for a reference model. 
The training objective of SimPO can be written as:
\begin{equation}
\mathcal{L}_{\mathrm{SimPO}}(\pi_{\theta}) = -\mathbb{E}_{(x,y_{w},y_{l})}\left[\log\sigma\left(\frac{\beta}{|y_{w}|}\log \pi_{\theta}(y_{w}| x) - \frac{\beta}{|y_{l}|}\log \pi_{\theta}(y_{l}| x)- \gamma \right)\right],
\label{equ:simpo}
\end{equation}
where $\pi_{\theta}$ is the policy model~(model to be optimized), $\beta$~(scaling of the reward difference) and $\gamma$~(target reward margin) are the hyper-parameters to separate the preferred and dis-preferred responses.
 
\paragraph{Efficient Context Scaling with Positional Indices Synthesis}
\label{para:skip}
Transformer-based models rely on positional indices to identify the relative position of each token~\citep{raffel2020exploring}. 
One efficient method to extend the data context length is modifying the positional indices to simulate long-sequence inputs without altering the real input sequence~\citep{press2021train,ruoss2023randomized}. 
By default, the positional indices of a sequence of length $k$ are $\mathcal{P}(k) = \{0,1,\cdots,k-1\}$. 
To extend the sequence length from $k$ to $K$, we can synthesize the positional indices: $\mathcal{P}_{\mathcal{B}}(K) = \{0 + b_{0}, 1+ b_{1},\cdots,k-1 + b_{k-1}\}$, where $\mathcal{B}=\{b_{0}, b_{1}, \cdots, b_{k-1}\}$ is the positional bias applied to each original position index and $k-1 + b_{k-1} = K$. 
To ensure effectiveness, the synthesis of position indices should achieve a uniform distribution of relative distances within the extended sequence length $[0, K]$ and cover as many of the extended positional indices as possible~\citep{wu2024long}.
% Therefore, the positional bias $\mathcal{B}$ should be meticulously designed.

% \paragraph{Relative Position Embedding} To expand the context window of an LLM pre-trained with a context window size $L_{s}$, activating positional encoding for the target length $L_{t}$ is crucial. 
% The use of relative positional encoding~\citep{shaw2018self} is currently the most common method for extending the model context length due to its scalability, and RoPE~\citep{su2023enhanced} is the most frequently used one~\citep{touvron2023llama,qwen2}. Given position $m$ and $n$ within a sequence, RoPE utilizes the positional encoding function $f(\cdot)$ to embed the positional information of $m$ and $n$ into each query $q$ vector and key $k$ vector at each layer. Therefore, the attention score $a(q_{m}, k_{n})$ becomes:
% \begin{equation}
%  a(q_{m}, k_{n}) = \left \langle f(q_{m}, m), f(k_{n}, n) \right \rangle = f(q_{m}, m)^{T}f(k_{n}, n) = g(q_{m}, k_{n}, \theta, m-n),
% \label{equ:rope}
% \end{equation}
% where $\theta$ is the base value that controls the properties of length extrapolation in RoPE.

\subsection{Long-context alignment with LOGO}
\label{subsec:lao}

% To enhance the LCM's ability to leverage the retrieval information for generation and mitigate the misalignment phenomenon, we introduce LOGO---Long cOntext aliGnment Optimization, which contains two parts: training objective~(\cref{sebsub:mosimpo}) and dataset construction~(\cref{subsec:logo_data}).

\subsubsection{Training Objective of LOGO}
\label{sebsub:mosimpo}
In long-context scenarios, LCMs are prone to generating various misaligned responses, such as hallucinations and failing to follow instructions~\citep{belyi2024luna}.
However, there is a lack of effective strategies~(or models) to detect these misaligned outputs, posting a great challenge for selecting preference and dis-preference samples in preference optimization~(we will elucidate this in Appendix~\ref{appdix:design_logo_train_obj}, where we also show the misalignment cases).
Therefore, instead of finding one dis-preference instance with a specific error pattern, we can expand the dis-preference space to push the model away from a range of possible dis-preference instances.
We design the loss function based on SimPO~(Eq.~\ref{equ:simpo}), as it is more aligned with the generation tasks and free of the reference model, which is efficient for long-context training.
The training objective can be written as:
\begin{small}
\begin{equation}
\mathcal{L}_{\mathrm{LOGO}}(\pi_{\theta}) = -\mathbb{E}_{(x,y_{w},y_{l}^{(1\cdots M)})}\left[\log\sigma\left(\frac{\beta}{|y_{w}|}\log \pi_{\theta}(y_{w}| x) - \frac{\beta}{M|y_{l}|}\sum_{j=1}^{M}\log \pi_{\theta}(y_{l}^{(j)}| x)- \gamma \right)\right],
\label{equ:logo}
\end{equation}
\end{small}
where $M$ is the number of dis-preference instances.

Furthermore, to avoid reward hacking phenomenon~\citep{yuan2024rrhf,hong2024reference} as well as preserve the modeling capabilities of LCMs, we add an SFT regularization term in Equ~\ref{equ:logo}.
This regularization term serves to prevent the policy model $\pi_{\theta}$ from drifting away from its original capabilities acquired through SFT. 
The final loss function of LOGO can be written as:
\begin{equation}
\mathcal{L}_{\mathrm{LOGO}}^{*}(\pi_{\theta}) = \mathcal{L}_{\mathrm{LOGO}}(\pi_{\theta}) + \lambda\mathbb{E}_{(x,y_{w})} \log \pi_{\theta}(y_{w}|x)),
\label{equ:logo_obj_reg}
\end{equation}
where $\lambda$ is the hyper-parameter that controls SFT regularization term.  

\subsubsection{Training Dataset Construction of LOGO}
\label{subsec:logo_data}
\begin{figure}[t]
    \centering
    \includegraphics[width=1\linewidth]{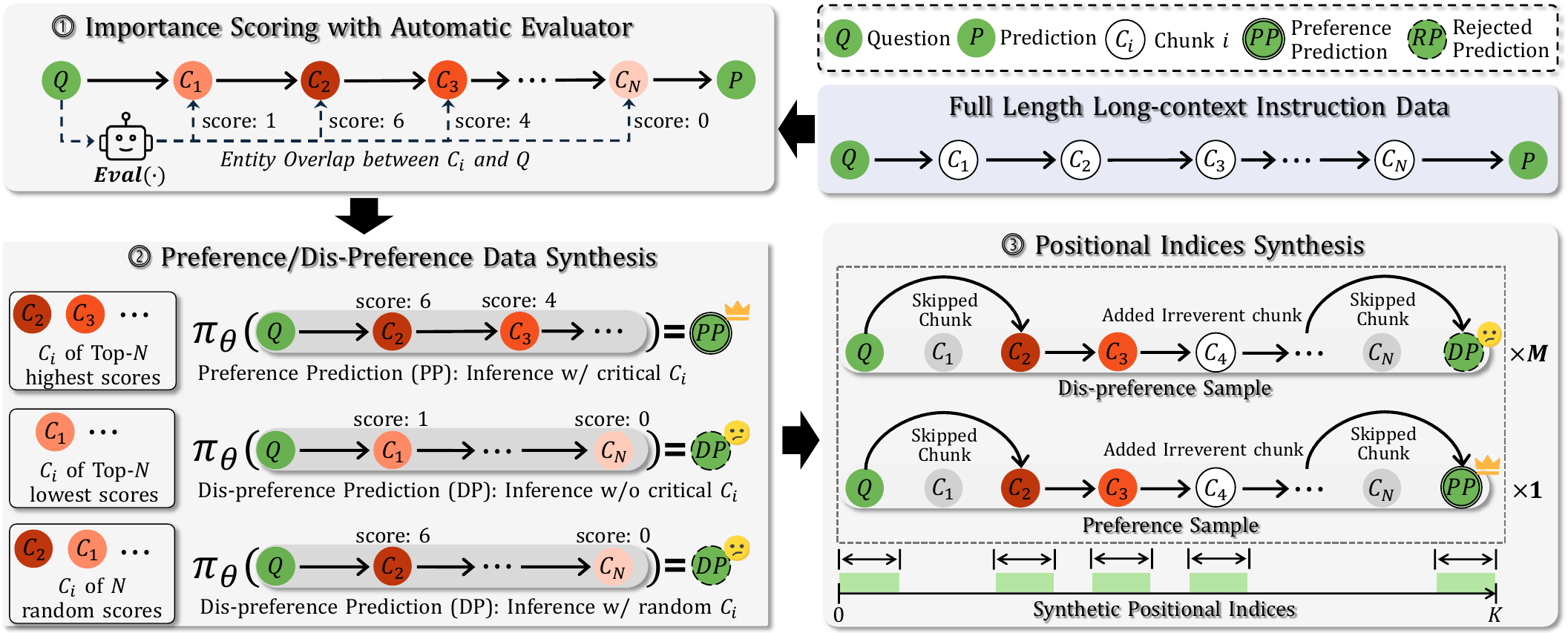}
    \caption{Dataset construction pipeline of LOGO.}
    \vspace{-1em}
    \label{fig:data_syn}
\end{figure}

To perform the LOGO training, we introduce a tailored LOGO dataset construction pipeline.
For each long-context sample, we can format it as a triplet $\mathcal{X} = \{Q, \mathcal{C}, P\}$, where $Q$, $\mathcal{C}$, and $P$ represent the question, reference context, and the model prediction, respectively. 
As shown in Fig.~\ref{fig:data_syn}, to construct training data for LOGO, we first divide the context $\mathcal{C}$ into equal-length chunks $\{C_1, C_2, \cdots, C_n\}$.
Then, three steps are involved: (1) Importance Scoring with Automatic Evaluator, (2) Preference and Dis-preference Data Synthesis, and (3) Positional Indices Synthesis.

\paragraph{Importance Scoring with Automatic Evaluator}
To construct preference~(aligned) and dis-preference~(misaligned) data in long-context scenarios, an efficient method is to guide the model to respond based on different contexts.
Specifically, to construct the preference data, we only provide the model with context relevant to the question, thus enhancing the fidelity of the model's output by reducing contextual interference~\citep{shi2023large}. 
Conversely, we can add more irrelevant context to guide the model in generating misaligned content like hallucinations.
To find the relevant chunks~$C_i$ within the context, we utilize an automatic evaluator $\mathrm{Eval}(\cdot)$ to calculate the ``contribution'' of each chunk $C_i$ to the question $Q$.
Specifically, we utilize an $\mathrm{Eval}(\cdot)$ to identify all the entities within a chunk $C_i$. 
The more overlapping entities $C_i$ shares with the question $Q$, the greater its influence on the final prediction, allowing us to assign a higher score to this chunk.
With $\mathrm{Eval}(\cdot)$, we efficiently assign importance scores $S=\{s_1, s_2, \cdots, s_n\}$ to all the chunks.

\paragraph{Preference and Dis-preference Data Synthesis}
To construct preference and dis-preference data based on the model prediction $P$, we select and combine the chunks mentioned above to create diverse contexts, guiding the model to generate different outputs.
Let $N$ represent the number of chunks within a context, and we define a threshold $\delta$ to distinguish between critical and irreverent chunks.
Specifically, chunks $\mathcal{C}_{>\delta}$ scoring above $\delta$ are considered as essential chunks while chunks $\mathcal{C}_{<\delta}$ scoring below $\delta$ are considered as irreverent chunks.
Then, we combine $Q$ and $\mathcal{C}_{>\delta}$ for model to generate preference prediction $P_{\mathrm{preference}}$, and adjust the ratio of chunks sampled from $\mathcal{C}_{>\delta}$ and $\mathcal{C}_{<\delta}$ for model to generate dis-preference predictions $P_{\mathrm{dis-preference}}$.
Specifically, $P_{\mathrm{dis-preference}}$ is mainly sampled from two misaligned error patterns: (1) model generation based on all irrelevant chunks $P_{\mathrm{dis-preference}}^{\prime}$, and (2) model generation based on partially relevant chunks $P_{\mathrm{dis-preference}}^{\prime\prime}$.
The above data construction process can be written as:
\begin{equation} \nonumber
\left\{
    \begin{aligned}
        & P_{\mathrm{preference}} =  \pi_{\theta}(Q, \mathcal{C}_{>\delta})~\mathrm{,where}~~\mathcal{C}_{>\delta} \sim \mathcal{C}, |\mathcal{C}_{>\delta}|=N \\
        & P_{\mathrm{dis-preference}} \sim \left\{
        \begin{aligned}
        & P_{\mathrm{dis-preference}}^{\prime} =  \pi_{\theta}(Q, \mathcal{C}_{<\delta})~\mathrm{,where}~~\mathcal{C}_{<\delta} \sim \mathcal{C}, |\mathcal{C}_{<\delta}|=N~,\\
        & P_{\mathrm{dis-preference}}^{\prime\prime} = \pi_{\theta}(Q, \mathcal{C}_{<\delta}, \mathcal{C}_{>\delta})~\mathrm{,where}~~\mathcal{C}_{<\delta}, \mathcal{C}_{>\delta} \sim \mathcal{C}, |\mathcal{C}_{<\delta} \cup \mathcal{C}_{>\delta}|=N 
        \end{aligned}
        \right\}
    \end{aligned}
    \right..
\end{equation}

Subsequently, the constructed preference and dis-preference data share the same context $\mathcal{C}^{\prime}$, which is combined with all the chunks in $\mathcal{C}_{>\delta}$ and partial chunks in $\mathcal{C}_{<\delta}$. Finally, one LOGO training sample can be written as $\left(\{Q, \mathcal{C}^{\prime}, T_{\mathrm{preference}}\}, \{Q, \mathcal{C}^{\prime}, T^{(i)}_{\mathrm{dis-preference}}\}_{i=1}^{M} \right)$, which is consistent with Eq.~\ref{equ:logo_obj_reg}.

\paragraph{Positional Indices Synthesis}
Given that each LOGO training sample includes $(M+1)$ instances, with one preference instance and $M$ dis-preference instance, a long context length of $\mathcal{C}^{\prime}$ can easily lead to GPU memory overflow (even on GPUs with 80GB memory).
To address this, we employ a positional encoding synthesis strategy. 
By assigning different synthetic positional indices to each chunk, we can simulate long-sequence training data with short context data~\citep{wu2024long}.
Specifically, to ensure that the synthetic positional indices do not disrupt the semantic structure of short context, the positional indices within each chunk should be continuous, while indices between adjacent chunks can be discrete, i.e., omitting certain positional indices (as shown in sub-Fig.~\ding{174} in Fig.~\ref{fig:data_syn}).
Given $N$ equal-length chunks within each sample\footnote{Since the length of question $\mathcal{Q}$ and prediction $P$ are much shorter compared to the long context $\mathcal{C}$, we can ignore the length of $\mathcal{Q}$ and $\mathcal{P}$ for simplicity.}, to achieve a uniform distribution of relative distance within the expanded context length $[0, K]$, each positional bias term $b_i \in \mathcal{B}$ should be sampled from a uniform distribution. 
The synthetic positional indices can be written as:
\begin{equation}
\mathcal{P}_{\mathcal{B}}(K) = \{i + b_i\}_{i=0}^{k-1},~\mathrm{where}~~b_i \sim \mathcal{U}(1, (i\bmod |C_{i}|) \times (K-k)/N),
\label{equ:pos_syn}
\end{equation}
where $\left(i\bmod |C_{i}|\right)$ indicates the chunk index where the current positional index $i$ resides, and $(K-k)/N$ represents the expansion size for each chunk. More details are shown in Appendix~\ref{appdix:pos_synthesis}.
\section{Experiment}
\label{sec:experiment}

\subsection{Settings}
\label{subsec:settings_exp}
\paragraph{LOGO Dataset Construction}
We construct the LOGO datasets based on two corpora: (1) 4,000 instances sampled from long-llm-data\footnote{\url{https://huggingface.co/datasets/namespace-Pt/long-llm-data}}~\citep{zhang2024extending}, which includes reference contexts from multiple domains~(e.g., biography, paper, \textit{etc.}) and questions generated by GPT-4, covering tasks such as Single-Detail QA, Multi-Detail QA, and Summarization; (2) 2,000 instances sampled from RedPajama~\citep{together2023redpajama} to mitigate forgetting, where we prompt the open-source LCM Qwen2-70B-Instruct~\citep{qwen2} to generate questions for each instance. 
Then, we split each instance into equal-length chunks, with each chunk containing 512 tokens.
To construct preference and dis-preference data, we use the spaCy model\footnote{\url{https://spacy.io/usage/models}}, a named entity recognition~(NER) model that can identify all the entities within a context, as the evaluator~$\mathrm{Eval(\cdot)}$. 
We use the number of overlapping entities between each chunk $C_i$ and the question $Q$ as the importance score.
We set the threshold $\delta$ as 6, and chunk number $N$ as 16, i.e., selecting and combining 16 chunks as the reference context for training.
As for the number of dis-preference instances in the LOGO training objective, we set $M = 2$, i.e., each training sample includes one preference instance and two dis-preference instances.
Then, we apply Eq.~\ref{equ:pos_syn} to construct positional indices for each instance within each sample. Specifically, we adopt two different sampling strategies on positional bias $\mathcal{B}$ to ensure that all positional indices are uniformly covered and maintain the semantic structure of the context~(see Appendix~\ref{appdix:pos_synthesis} for more details). 
After positional indices synthesis, we have a total number of 12,000 training samples, with a total data size of approximately 12,000$\times$512$\times$16$\times$3$\approx$0.3B tokens.

\paragraph{Training Settings}
To improve the training efficiency while preserving the inherent capabilities of the LLMs, we freeze the backbone model and apply LoRA~\citep{hu2021lora} method, which only fine-tunes the attention and token embedding modules, to perform training.
Additionally, thanks to positional indices synthesis, LOGO can potentially scale the context length and ensure alignment in long-context tasks simultaneously.
Therefore, we experiment with two type of models: (1) Short-context Models~(SCMs) including Llama-2-7B-Chat~\citep{touvron2023llama} and Llama-3-8B-Instruct~\citep{llama3modelcard}, which own context lengths of 4K and 8K, respectively; and (2) Long-context Models~(LCMs), including Llama3-8B-Instruct-80K~\citep{zhang2024extending}, Llama-2-7B-Instruct-80K~\citep{fu2024data} and Mistral-Instruct-7B-V0.2~\citep{jiang2023mistral}, which inherently have long context windows.
For SCMs, given that excessive scaling with positional indices synthesis method can result in the missing of some positional indices, potentially impacting model performance, we scale the context windows of SCMs to 8 times of their original context length.
For LCMs, we maintain their original context length.
To accelerate the training process and save GPU memory, we adopt DeepSpeed Zero 3~\citep{aminabadi2022deepspeed}.
All the experiments are conducted on a 8$\times$A800~(80GB) GPU machine, and the training is completed within 16 hours.
For the setting of hyper-parameters $\beta$ and $\gamma$ in Eq.~\ref{equ:logo}, we adhere to the recommendations provided in~\citet{meng2024simpo} for different models, where $\beta=10, \gamma=3$ for Llama-3-8B-based model, $\beta=2.5, \gamma=0.25$ for Mistral-Instruct-7B-V0.2-based model, and $\beta=3, \gamma=0.6$ for Llama-2-7B-based model.
We set $\lambda=0.1$ in Eq.~\ref{equ:logo_obj_reg} for SFT regularization to stabilize the training process of LOGO and prevent the reward hacking phenomenon mentioned above.
% \vspace{-0.5em}
\paragraph{Evaluation Settings}
We assess the LOGO training strategy across three categories of long-context tasks: real-world long-context tasks, a synthetic retrieval task, and the language modeling task.
To explore the impact of LOGO training in short-context scenarios, we also evaluate models on short-context tasks.
For comparison, we select two representative context scaling methods: YaRN~\citep{peng2023yarn} and RandPOS~\citep{ruoss2023randomized}, as well as two types of long-instruction tuning strategies~\cite{xiong2023effective}, i.e., calculating loss on the entire sequence~(Full) and the prediction~(Partial).
We select LongAlpaca~\citep{long-alpaca} corpus as the instruction training data, which contains 12,000 long instruction samples with each sample containing 32K context length.

\subsection{Performance on Long-context Tasks}
\paragraph{Results on Real-world Long-context Tasks}
\begin{table*}[tb]
\centering
\small
\caption{Evaluation results on LongBench benchmark, where $\dagger$ denotes training-free method.}
\vspace{-0.5em}
% \resizebox{\linewidth}{!}{
\begin{tabular}{l c c c c c c}
\toprule
\textbf{Models} & \textbf{S-Doc QA} & \textbf{M-Doc QA} & \textbf{Summ} & \textbf{Few-shot} & \textbf{Synthetic} & \textbf{Avg.} \\
\midrule
GPT-3.5-Turbo-16K & 39.8 & 38.7 & 26.5 & 67.1 & 37.8 & 42.0 \\
LongChat-v1.5-7B-32k & 28.7 & 20.6 & 26.7 & 60.0 & 15.8 & 30.4 \\
% ChatGLM3-6B-32K & 40.3 & 46.6 & 29.5 & 68.1 & 50.5 & 47.0 \\
LLama-3.1-8B-Instruct-128K & 23.9 & 15.8 & 28.9 & 69.8 & 57.5 & 39.2 \\
\midrule
\rowcolor{yellow} \multicolumn{7}{c}{\textit{Results on SCMs (scaling $\times$8 context window)}} \\
\midrule
Llama-3-8B-Instruct-8K & 39.3 & 36.2 & 24.8 & 63.5 & 39.9 & 40.7 \\
~~~+ ~YaRN-64K$^{\dagger}$ & 38.0 & 36.6 & 27.4 & 61.7 & 40.9 & 40.9 \\
~~~+ ~RandPOS-64K & 32.5 & 30.5 & 26.5 & 61.3 & 33.4 & 36.8 \\
~~~+ ~LOGO-64K & \bf 39.8 & \bf 36.7 & \bf 28.8 &\bf  65.4 &\bf  49.0 &\bf  43.9 \\
\cmidrule{1-7}
Llama-2-7B-Chat-4K & 24.9 & 22.6 & 24.7 & 60.0 & 5.9 & 27.6\\
% ~~~+ ~YaRN-32K~$^{\dagger}$ & \\
% ~~~+ ~RandPOS-32K \\
~~~+ ~LOGO-32K &\bf  26.7 & \bf 23.3 & \bf 26.3 & \bf 63.1 & \bf 11.1 &\bf  30.1 \\
\midrule
\rowcolor{cyan!25}\multicolumn{7}{c}{\textit{Results on LCMs~(long-context alignment)}} \\
\midrule
 Llama-3-8B-Instruct-80K & 43.0 & 39.8 & 22.2 & 64.3 & 46.3 & 42.3 \\ 
~~~+ ~Instruct Tuning~(Full)  & 38.8 & 35.0 & 24.6 & 65.9 & 44.5 & 41.8 \\ 
~~~+ ~Instruct Tuning~(Partial) & 39.3 & 36.2 & 26.8 & 63.5 & 48.0 & 42.8 \\ 
~~~+ ~LOGO-80K & \bf 44.0 & \bf 41.2 &\bf  28.1 & \bf 68.6 & \bf 53.0 & \bf 47.0 \\ 
\cmidrule{1-7}
Llama-2-7B-Instruct-80K & 26.9 & 23.8 & 21.3 & 65.0 & 7.9 & 29.0 \\ 
~~~+ ~LOGO-80K &\bf  33.6 & \bf 28.0 & \bf 29.4 &\bf  65.1 &\bf 24.5 & \bf 36.1 \\
\cmidrule{1-7}
Mistral-Instruct-7B-V0.2-32K & 31.7 & 30.6 & 16.7 & 58.4 &  17.9 & 31.1 \\
~~~+ ~LOGO-32K &\bf  38.3 & \bf 37.6 & \bf 26.1 & \bf 67.0 & \bf 31.5 & \bf 40.1  \\
\bottomrule
\end{tabular}

\label{tab:longbench}
\end{table*}
We evaluate the LOGO performance with real-world long-context tasks in LongBench~\citep{bai2023longbench}, a comprehensive benchmark suite encompassing 16 distinct datasets spread across 6 task categories, including Single Document QA~(S-Doc QA), Multi-Document QA~(M-Doc QA), Summarization~(Summ), Few-shot, Synthetic, and Code. 
It is worth noting that we exclude the Code category since the code testing data primarily involves contexts of just around 4,000 tokens and our training data does not cover this domain.
We report the evaluation results in Tab.~\ref{tab:longbench}, where we can observe that: (1) \textbf{LOGO achieves the best performance among all the settings}. Specifically, for SCMs, LOGO outperforms both YaRN and RandPOS. Although these two methods can potentially extend the context window of SCMs, they significantly impair performance on real-world long-context tasks. For instance, RandPOS causes the Llama3-8B-Instruct model to drop around 6 points on average compared to the baseline, with particularly notable declines in performance on the synthetic tasks.
For LCMs, LOGO can significantly improve model performance, with all LCMs showing varying degrees of improvement, e.g., Llama-3-8B-Instruct-80K model shows an average 5-point improvement compared to the baseline, whereas the instruct tuning method tends to restrict even a well-performing LLMs to a limited performance bottleneck; 
(2) \textbf{Compared to other methods, LOGO demonstrates significant improvement in information-intensive tasks}, which require the model to gather information from various parts of the context. 
Specifically, in summarization and synthetic tasks, LCMs trained with LOGO can achieve significant performance improvements, with at least a 5-point increase.

\paragraph{Evaluation Results on Synthetic Retrieval Task}
\begin{figure}[t]
    \centering
    \includegraphics[width=1\linewidth]{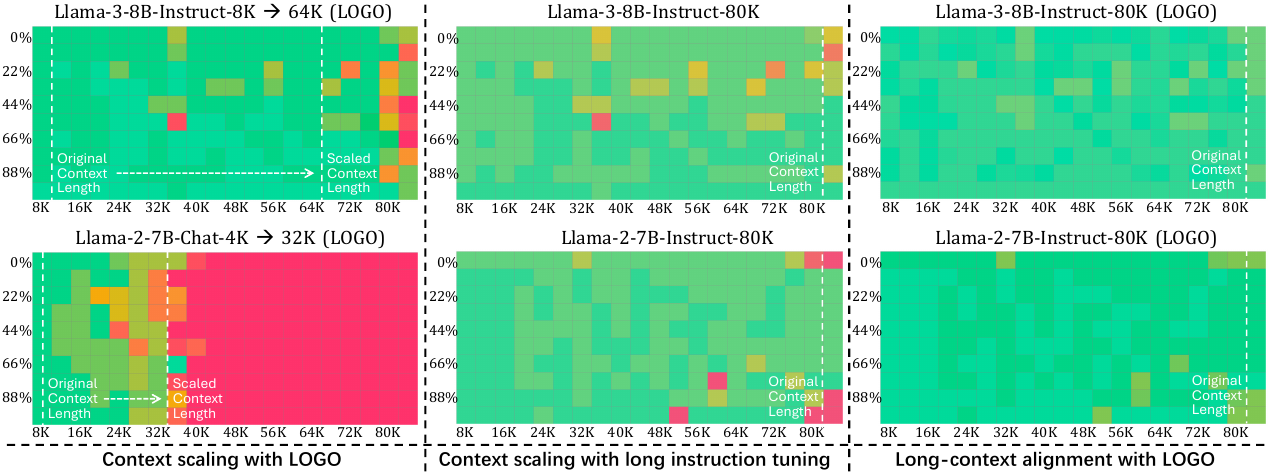}
    \vspace{-1.5em}
    \caption{Results of the Needle-in-a-Haystack testing.}
    \vspace{-1em}
    \label{fig:NIAH}
\end{figure}
To investigate whether the LOGO training strategy affects the information retrieval capabilities of LCMs, we conduct a Needle-in-a-Haystack testing~\citep{NIAH}.
More concretely, NIAH is a synthetic retrieval task that evaluates a model's ability to retrieve key information~(needle) from any position within its context window.
We employ a color scale ranging from light green~(indicating a 100\% successful recall), to red~(indicating a complete failure).
Our test covers context lengths from 8K to 88K, with intervals of 0.5K and the needle at various depths.
As shown in Fig.~\ref{fig:NIAH}, we can find that LOGO can scale the context window for SCMs~(left group) and does not adversely affect the original context window size of LCMs~(right group).
We can also observe that the original LCMs~(middle group) and those trained with LOGO~(right group) share similar patterns, i.e., similar shades of color, yet LOGO improves performance in areas where the original LCMs fail.
This indicates that LOGO does not compromise the inherent capabilities of LCMs but rather enhances their original weakness.
\begin{wrapfigure}{r}{0.5\textwidth}
\centering
\includegraphics[width=1\linewidth]{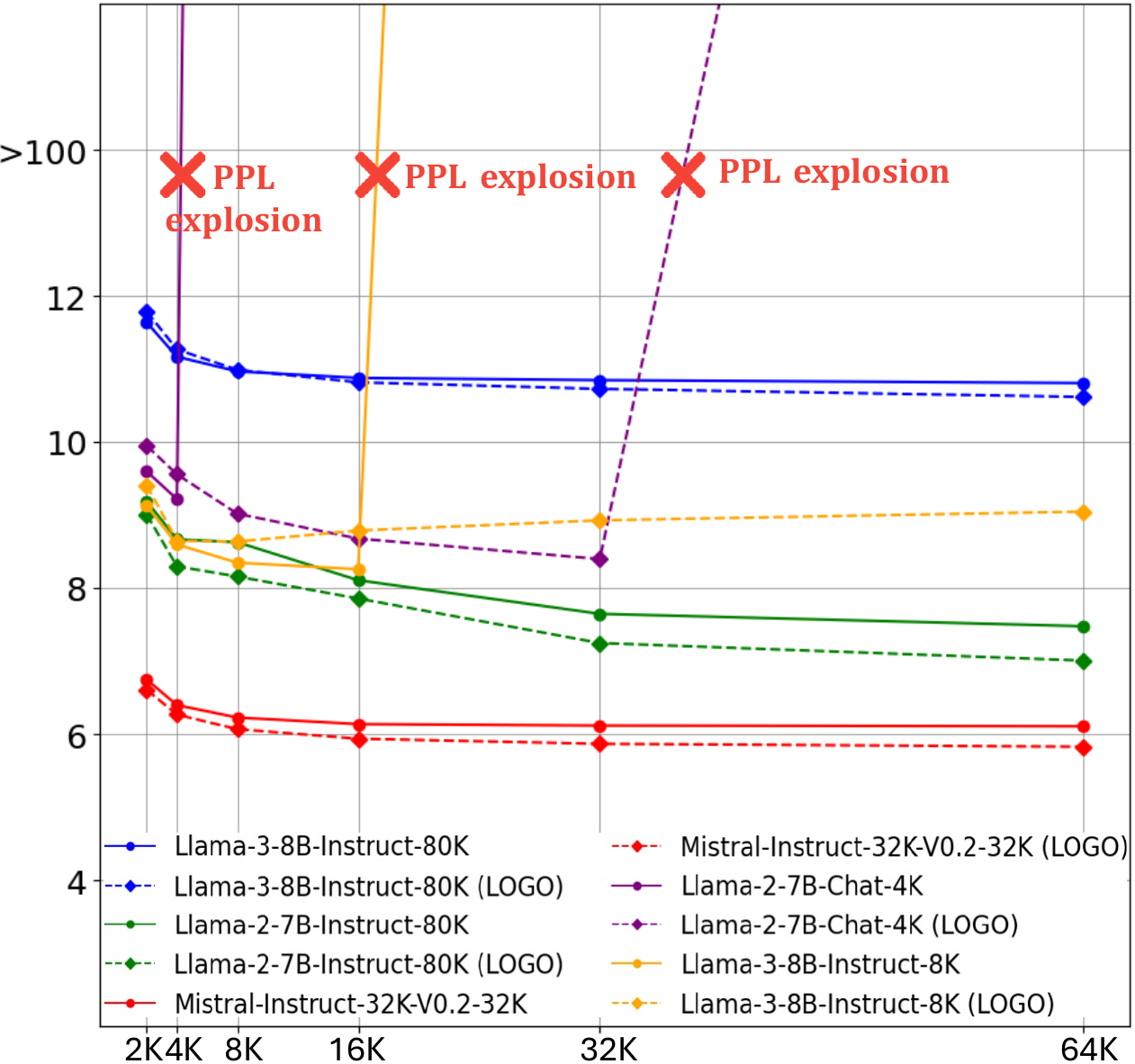}
\caption{Evaluation results of language modeling task. The solid and dashed curves represent the PPL of the baselines and LOGO, respectively.}
\label{fig:ppl}
\vspace{-2em}
\end{wrapfigure}
We can also find that the Llama-3-8B-8K model demonstrates superior context scaling effects compared to the Llama-2-7B-4K model. This can be attributed to the larger RoPE base value in Llama-3-8B-8K~(500,000) compared to Llama-2-7B-4K~(10,000), which has been proven to facilitate more effective scaling of the context window size~\citep{llama3modelcard}.

\paragraph{Evaluation Results on Language Modeling Task}
We test the language modeling capability of LCMs by calculating the Perplexity~(PPL) on the Gutenberg (PG-19) testing set~\citep{rae2019compressive}, with context lengths ranging from 2K to 64K.
Considering that extremely long context lengths can cause the PPL calculation to exceed GPU memory, we apply the sliding window approach proposed by \citet{press2021train}.
As depicted in Fig.~\ref{fig:ppl}, for LCMs, such as Llama-3-8B-Instruct-80K and Llama-2-7B-Instruct-80K, using LOGO does not compromise the language modeling capability since the solid line (PPL of the backbone model) and the dashed line (PPL of LOGO) almost completely overlap. 
In the case of SCMs, such as the Llama-3-8B-Instruct-8K model, LOGO not only effectively scales the context window size of baseline models~(the purple dotted curve versus the purple solid curve) but also achieves a lower PPL score compared to the SFT method since the yellow dotted curve~(PPL of Llama-3-8B-Instruct-LOGO) is much lower than the blue solid curve~(PPL of Llama-3-8B-Instruct-80K).

\subsection{Performance on Short-context Tasks}
\begin{figure}[t]
    \centering
    \includegraphics[width=1\linewidth]{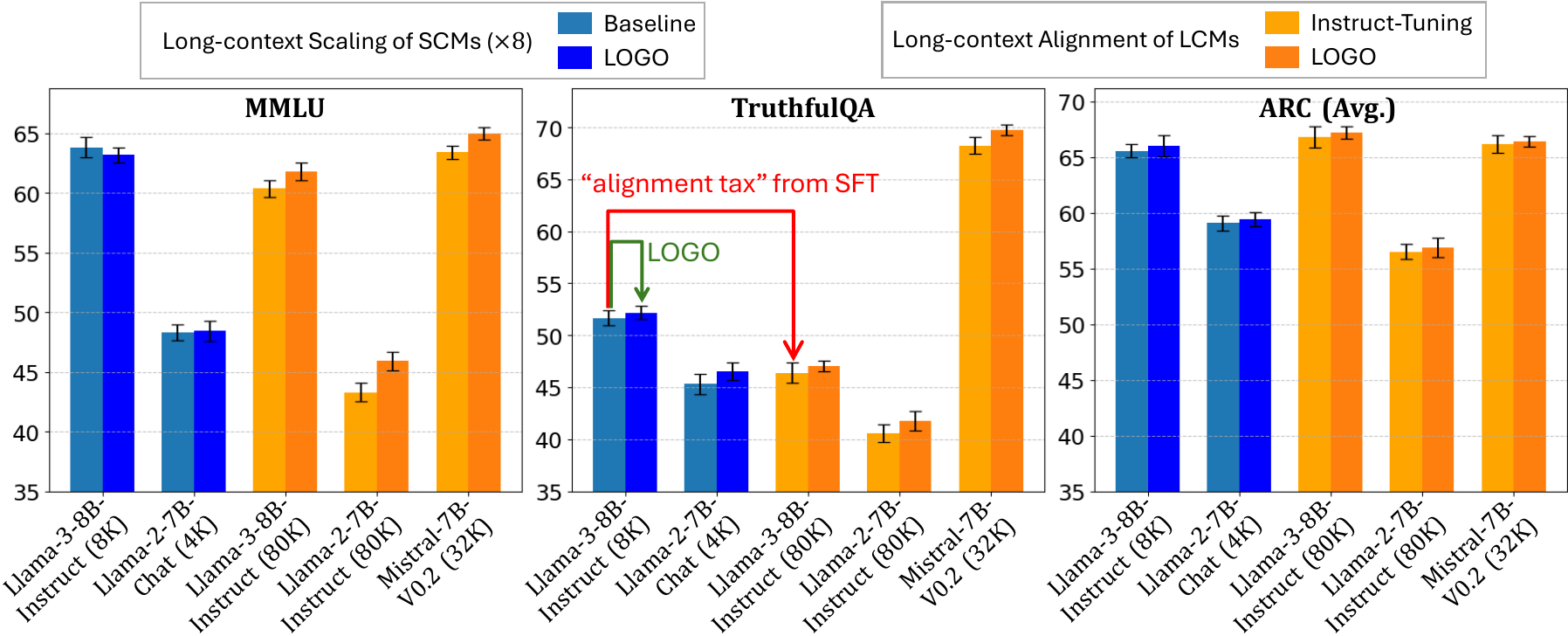}
    \vspace{-1.5em}
    \caption{Model performance on short-context tasks, including MMLU, TruthfulQA, and ARC.}
    \label{fig:short_ctx_tasks}
\end{figure}

To investigate whether LOGO training affects model performance on short-context tasks, we select three widely used benchmarks for assessing LLMs' foundational capabilities that possess short input sequence: MMLU~\citep{hendrycks2020measuring}, TruthfulQA~\citep{lin2021truthfulqa}, and ARC~(Hard and Easy)~\citep{clark2018think}.
As illustrated in Fig.~\ref{fig:short_ctx_tasks}, we find that LOGO not only preserves the LLM's inherent capabilities on short-context tasks but also demonstrates improvements in some specific tasks.
This is because LOGO aims to teach the model to generate responses based on the context rather than fabricating results~(such as producing hallucinations), which is equally applicable to short-context tasks.
We can also find that scaling context length with LOGO yields better results than instruction tuning.
For instance, as demonstrated in the TruthfulQA task, Llama-3-8B-Instruct-80K shows significant performance degradation compared to the Llama-3-8B-Instruct-8K-LOGO~(64K).
Such a phenomenon indicates a high ``alignment tax'' paid from instruction tuning~\citep{fu2023chain}.

\section{Ablation Study}
\label{sec:ablation_study}
For ablation studies, we experiment with the Llama-3-8B-Instruct-80K model, which demonstrates strong baseline performance across the various tasks.
We conduct experiments on the real-world tasks by reporting the average score on LongBench~(denoted with $\mathrm{LB}$), and the language modeling task by calculating the PPL score on the PG-19 testing set with a 64K context length.
In Sec.~\ref{subsec:ana_logo_obj}, we analyze the impact of different hyper-parameters in the LOGO training objective. 
In Sec.~\ref{subsec:effect_syn_data}, we discuss the impact of synthetic data of varying lengths.
In Sec.~\ref{subsec:com_logo_sft}, we compare LOGO with SFT by visualizing LCM's generation and information retrieval capabilities along the training phase.

\begin{figure}[t]
    \centering
    \includegraphics[width=\linewidth]{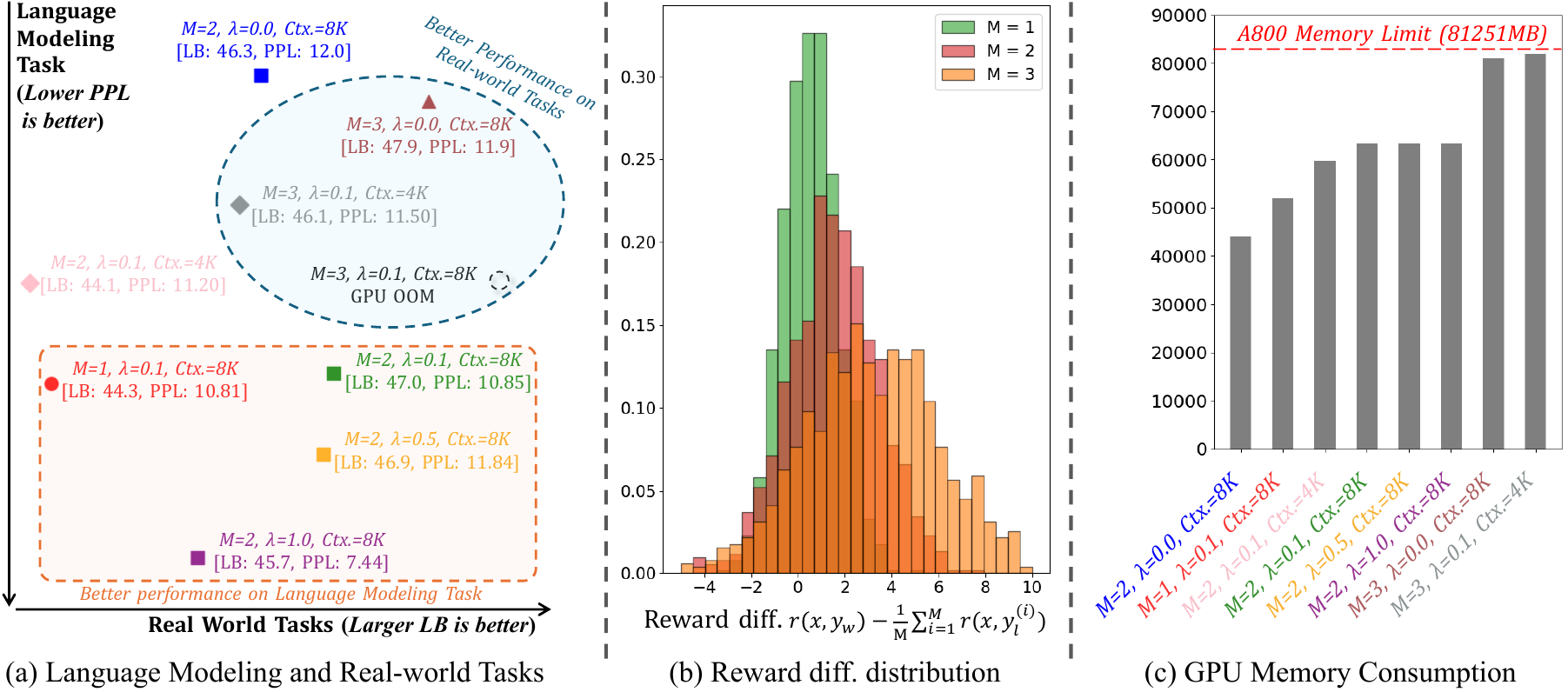}
    \caption{Ablation study results. (a) Comparison among different settings on the language modeling task~(PPL) and real-world tasks~(Avg.~score on LongBench testing set); (b) Reward difference distribution under different $M$ settings; (c) Training GPU memory consumption of different settings.}
    \label{fig:ablation_studies}
    \vspace{-0.5em}
\end{figure}

\subsection{Analysis of LOGO Training Objective}
\label{subsec:ana_logo_obj}
% To clearly present the comparisons, we show the performance across different settings in Fig.~\ref{fig:ablation_studies}, which includes model performance, reward margin of LOGO training objective, as well as reporting the GPU memory consumption during training for different configurations (including the regularization term $\lambda$, the number of dis-preference instances $M$, and the actual input context length $Ctx.$).
\paragraph{Effect of SFT Regularization Term $\lambda$}
To investigate the SFT regularization term in Equ.~\ref{equ:logo_obj_reg}, we adjust the value of $\lambda$ to control the SFT regularization term.
As depicted in Fig.~\ref{fig:ablation_studies}(a), we can observe that increasing $\lambda$ enables the model to achieve a lower PPL score.
For real-world tasks, the impact of SFT regularization on the final results is minimal.
For example, for settings $(M=2, \lambda=0.1, Ctx.=8K)$, $(M=2, \lambda=0.5, Ctx.=8K)$, and $(M=2, \lambda=1.0, Ctx.=8K)$, we can observe that as $\lambda$ gradually increases, the PPL significantly decreases, with a difference of nearly 3.5 points, while the average score on LongBench only differs by around 1.5 points.

\begin{figure}[t]
    \centering
    \includegraphics[width=\linewidth]{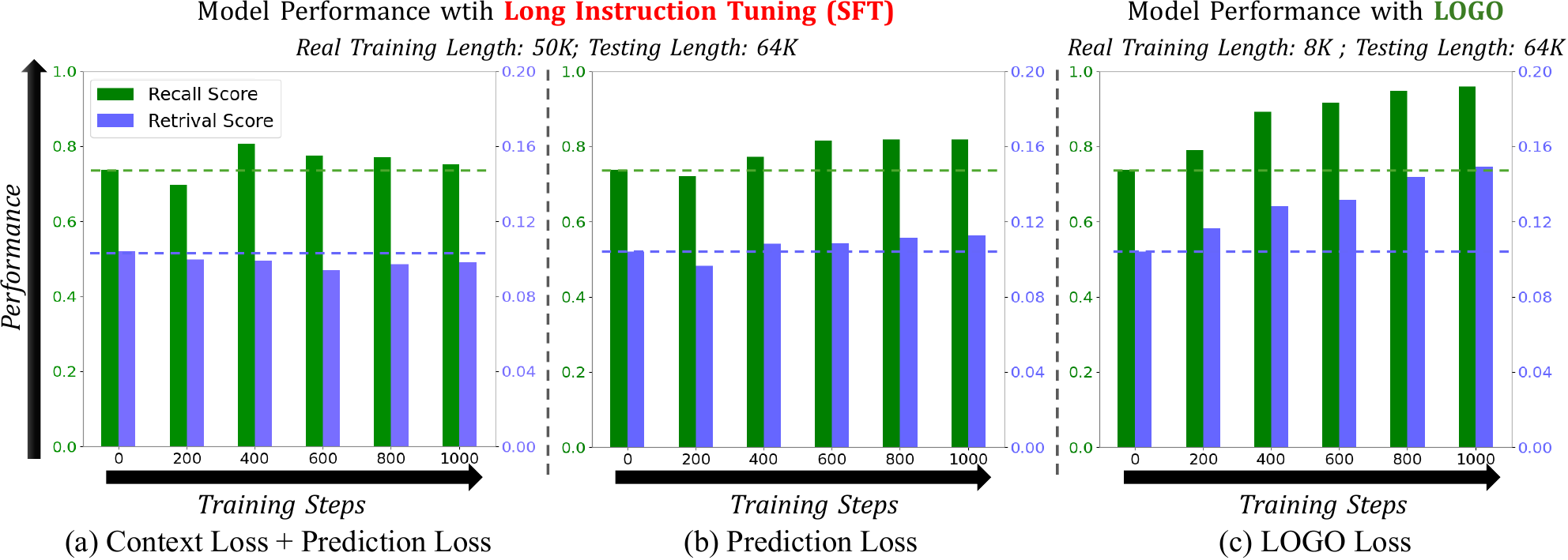}
    \caption{Comparison between SFT and LOGO training strategies on the synthetic retrieval task.}
    \label{fig:retrival_head}
    \vspace{-1em}
\end{figure}

\paragraph{Effect of the Number of Dis-Preference Instances}
We experiment with different numbers of dis-preference instance $M = \{1, 2, 3\}$ in Eq.~\ref{equ:logo_obj_reg}. Specifically, when $M$ equals 1, the LOGO Objective degenerates into the SimPO Objective. 
As shown in Fig.~\ref{fig:ablation_studies}(a), using more dis-preference samples can enhance the model's performance on real-world tasks, but it slightly impacts the capability for language modeling.
We also visualize the learned reward margin $r(x, y_w) - \frac{1}{M}\sum_{i=1}^{M}r(x, y_{l}^{(i)})$ under various $M$ values in Fig.~\ref{fig:ablation_studies}(b). 
We can observe that using a larger $M$ can flatten the distribution and make it easier for the model to distinguish between preference and dis-preference samples as the gap between $r(x, y_w)$ and $\frac{1}{M}\sum_{i=1}^{M}r(x, y_{l}^{(i)})$ gradually increases with larger $M$.
This is because increasing \( M \) can cover more samples with various types of misalignment patterns.
However, as shown in Fig.~\ref{fig:ablation_studies}(c), increasing \( M \) poses a challenge as it may exceed GPU memory limits.
While introducing more dis-preference samples in the LOGO objective function might be beneficial, optimizing this in practical deployment is necessary. 
Additionally, the impact of each dis-preference sample's weight needs to be explored, which we will address in our further work.

\subsection{Effect of Synthetic Data Length}
\label{subsec:effect_syn_data}
We study with two settings of synthetic data length, i.e., from real input length 4K to target length 64K~($Ctx.=4K$) and from real input length 8K to target length 64K~($Ctx.=8K$). 
Specifically, the chunk size $|\mathcal{C}_{i}|$ remains unchanged, while we set the number of chunks as 8 and 16 for the above two settings, respectively.
As shown in Fig.~\ref{fig:ablation_studies}(a), short-context synthetic data length significantly diminishes the model's performance on both the language modeling task and real-world tasks~(data point $(M=2, \lambda=0.1, \mathit{Ctx.}=4K)$ versus data point $(M=2, \lambda=0.1, \mathit{Ctx.}=8K)$), but can still overcome the instruction tuning method~(42.8 average score on LongBench) and effectively reduces the GPU memory requirement during training~(Fig.~\ref{fig:ablation_studies}(c)).
This is because when the original context length is relatively small~(4K), it requires scaling up by a larger factor~(16 times) to reach the desired context length~(64K). During the positional indices synthesis process, some positional indices may miss or be infrequently activated, thereby impacting performance.

\subsection{Comparison between SFT and LOGO}
\label{subsec:com_logo_sft}
As shown in Fig.~\ref{fig:retrival_head}, we illustrate the impact of SFT (with two loss calculation strategies following~\citep{xiong2023effective}) and LOGO on the model's generation and understanding performance throughout the training process.
We plot the trends of retrieval score~(understanding ability) and recall score~(generation ability) along the training progress.
We can observe that applying SFT loss to the entire sequence leads to a gradual decline in the LCM's understanding ability, accompanied by performance fluctuations; while applying SFT loss solely to the prediction portion shows no significant improvement in model performance.
Nevertheless, applying LOGO can steer LCMs away from misaligned samples, thereby enhancing the recall score. 
Simultaneously, it improves comprehension abilities, enabling the model to retrieve more key information within the context.

\vspace{-0.5em}
\section{Conclusion}
\vspace{-0.5em}
\label{sec:conclusion}
In this paper, we find that commonly used training approaches for LCMs may degrade the model's generation capabilities, leading to misaligned outputs, such as hallucinations and instruction unfollowing.
To mitigate this issue, we introduce LOGO, a novel preference optimization training strategy for long-context alignment. 
Specifically, LOGO has two key components: (1) a reference-free preference optimization objective that teaches the model to distinguish between the preference and the dis-preference predictions, and (2) a data construction pipeline tailored for the training objective, both of which are designed to ensure the training efficiency and effectiveness.
By performing LOGO training on a single 8$\times$A800 GPU machine within 16 hours, LCMs can achieve great improvements in long-context tasks while maintaining their inherent capabilities.
Besides, LOGO can also potentially scale the context length of short-context models and achieve better generation performance compared to other frequently used context scaling methods.

% \section*{Reproducibility Statement}
% Based on the policy of ICLR-2025 Author Guide~\footnote{\url{https://iclr.cc/Conferences/2025/AuthorGuide}}, this Reproducibility Statement \textbf{\textit{does not count toward the page limit}} and will briefly describe the key algorithms presented in the paper for reproducibility.
% The LOGO consists of two key components: (1) \textbf{LOGO training objective}, where we introduce the algorithm in Sec.~\ref{sebsub:mosimpo}; provide the hyper-parameters in Sec.~\ref{subsec:settings_exp}; analyze the influence of hyper-parameters in Sec.~\ref{subsec:ana_logo_obj}; show the constructed cases in Appendix~\ref{appdix:case_study}; and (2) \textbf{LOGO Data Construction}, where we introduce the algorithm in Sec.~\ref{subsec:logo_data}; provide the hyper-parameters in Sec.~\ref{subsec:settings_exp} and Appendix~\ref{appdix:pos_synthesis}; and analyze the impact of hyper-parameters in Sec.~\ref{subsec:effect_syn_data}.
% For the preliminary experiments in Introduction~(Sec.~\ref{sec:intro}), we provide details in Appendix~\ref{appdix:into}.
% The code for this paper can be found in the anonymous code file submitted in Supplementary Material.

% \section*{Ethics Statement}
% This research is done in alignment with Microsoft’s responsible AI principles.

% \section*{Acknowledgement}
% We would like to express our gratitude to Microsoft Research Asia for its support in terms of computational resources.

\bibliography{main}

\begin{thebibliography}{53}
\providecommand{\natexlab}[1]{#1}
\providecommand{\url}[1]{\texttt{#1}}
\expandafter\ifx\csname urlstyle\endcsname\relax
  \providecommand{\doi}[1]{doi: #1}\else
  \providecommand{\doi}{doi: \begingroup \urlstyle{rm}\Url}\fi

\bibitem[Abdin et~al.(2024)Abdin, Jacobs, Awan, Aneja, Awadallah, Awadalla, Bach, Bahree, Bakhtiari, Behl, et~al.]{abdin2024phi}
Marah Abdin, Sam~Ade Jacobs, Ammar~Ahmad Awan, Jyoti Aneja, Ahmed Awadallah, Hany Awadalla, Nguyen Bach, Amit Bahree, Arash Bakhtiari, Harkirat Behl, et~al.
\newblock Phi-3 technical report: A highly capable language model locally on your phone.
\newblock \emph{arXiv preprint arXiv:2404.14219}, 2024.

\bibitem[Achiam et~al.(2023)Achiam, Adler, Agarwal, Ahmad, Akkaya, Aleman, Almeida, Altenschmidt, Altman, Anadkat, et~al.]{achiam2023gpt}
Josh Achiam, Steven Adler, Sandhini Agarwal, Lama Ahmad, Ilge Akkaya, Florencia~Leoni Aleman, Diogo Almeida, Janko Altenschmidt, Sam Altman, Shyamal Anadkat, et~al.
\newblock Gpt-4 technical report.
\newblock \emph{arXiv preprint arXiv:2303.08774}, 2023.

\bibitem[AI@Meta(2024{\natexlab{a}})]{llama3_1modelcard}
AI@Meta.
\newblock Llama 3-1 model card.
\newblock \emph{Blob}, 2024{\natexlab{a}}.
\newblock URL \url{https://ai.meta.com/blog/meta-llama-3-1/}.

\bibitem[AI@Meta(2024{\natexlab{b}})]{llama3modelcard}
AI@Meta.
\newblock Llama 3 model card.
\newblock \emph{Blob}, 2024{\natexlab{b}}.
\newblock URL \url{https://github.com/meta-llama/llama3/blob/main/MODEL_CARD.md}.

\bibitem[Aminabadi et~al.(2022)Aminabadi, Rajbhandari, Awan, Li, Li, Zheng, Ruwase, Smith, Zhang, Rasley, et~al.]{aminabadi2022deepspeed}
Reza~Yazdani Aminabadi, Samyam Rajbhandari, Ammar~Ahmad Awan, Cheng Li, Du~Li, Elton Zheng, Olatunji Ruwase, Shaden Smith, Minjia Zhang, Jeff Rasley, et~al.
\newblock Deepspeed-inference: enabling efficient inference of transformer models at unprecedented scale.
\newblock In \emph{SC22: International Conference for High Performance Computing, Networking, Storage and Analysis}, pp.\  1--15. IEEE, 2022.

\bibitem[anthropic(2024)]{claude_3_5}
anthropic.
\newblock Claude-3-5-sonnet model card.
\newblock \emph{blog}, 2024.
\newblock URL \url{https://www.anthropic.com/news/claude-3-5-sonnet}.

\bibitem[Bai et~al.(2023)Bai, Lv, Zhang, Lyu, Tang, Huang, Du, Liu, Zeng, Hou, Dong, Tang, and Li]{bai2023longbench}
Yushi Bai, Xin Lv, Jiajie Zhang, Hongchang Lyu, Jiankai Tang, Zhidian Huang, Zhengxiao Du, Xiao Liu, Aohan Zeng, Lei Hou, Yuxiao Dong, Jie Tang, and Juanzi Li.
\newblock Longbench: A bilingual, multitask benchmark for long context understanding.
\newblock \emph{arXiv preprint arXiv:2308.14508}, 2023.

\bibitem[Bai et~al.(2024)Bai, Lv, Zhang, He, Qi, Hou, Tang, Dong, and Li]{bai2024longalign}
Yushi Bai, Xin Lv, Jiajie Zhang, Yuze He, Ji~Qi, Lei Hou, Jie Tang, Yuxiao Dong, and Juanzi Li.
\newblock Longalign: A recipe for long context alignment of large language models.
\newblock \emph{arXiv preprint arXiv:2401.18058}, 2024.

\bibitem[Belyi et~al.(2024)Belyi, Friel, Shao, and Sanyal]{belyi2024luna}
Masha Belyi, Robert Friel, Shuai Shao, and Atindriyo Sanyal.
\newblock Luna: An evaluation foundation model to catch language model hallucinations with high accuracy and low cost.
\newblock \emph{arXiv preprint arXiv:2406.00975}, 2024.

\bibitem[Chen et~al.(2023{\natexlab{a}})Chen, Wong, Chen, and Tian]{chen2023extending}
Shouyuan Chen, Sherman Wong, Liangjian Chen, and Yuandong Tian.
\newblock Extending context window of large language models via positional interpolation.
\newblock \emph{arXiv preprint arXiv:2306.15595}, 2023{\natexlab{a}}.

\bibitem[Chen et~al.(2023{\natexlab{b}})Chen, Qian, Tang, Lai, Liu, Han, and Jia]{chen2023longlora}
Yukang Chen, Shengju Qian, Haotian Tang, Xin Lai, Zhijian Liu, Song Han, and Jiaya Jia.
\newblock Longlora: Efficient fine-tuning of long-context large language models.
\newblock \emph{arXiv preprint arXiv:2309.12307}, 2023{\natexlab{b}}.

\bibitem[Chen et~al.(2023{\natexlab{c}})Chen, Yu, Qian, Tang, Lai, Liu, Han, and Jia]{long-alpaca}
Yukang Chen, Shaozuo Yu, Shengju Qian, Haotian Tang, Xin Lai, Zhijian Liu, Song Han, and Jiaya Jia.
\newblock Long alpaca: Long-context instruction-following models.
\newblock \url{https://github.com/dvlab-research/LongLoRA}, 2023{\natexlab{c}}.

\bibitem[Clark et~al.(2018)Clark, Cowhey, Etzioni, Khot, Sabharwal, Schoenick, and Tafjord]{clark2018think}
Peter Clark, Isaac Cowhey, Oren Etzioni, Tushar Khot, Ashish Sabharwal, Carissa Schoenick, and Oyvind Tafjord.
\newblock Think you have solved question answering? try arc, the ai2 reasoning challenge.
\newblock \emph{arXiv preprint arXiv:1803.05457}, 2018.

\bibitem[Computer(2023)]{together2023redpajama}
Together Computer.
\newblock Redpajama: an open dataset for training large language models, 2023.
\newblock URL \url{https://github.com/togethercomputer/RedPajama-Data}.

\bibitem[Dao(2023)]{dao2023flashattention}
Tri Dao.
\newblock Flashattention-2: Faster attention with better parallelism and work partitioning.
\newblock \emph{arXiv preprint arXiv:2307.08691}, 2023.

\bibitem[Dubey et~al.(2024)Dubey, Jauhri, Pandey, Kadian, Al-Dahle, Letman, Mathur, Schelten, Yang, Fan, et~al.]{dubey2024llama}
Abhimanyu Dubey, Abhinav Jauhri, Abhinav Pandey, Abhishek Kadian, Ahmad Al-Dahle, Aiesha Letman, Akhil Mathur, Alan Schelten, Amy Yang, Angela Fan, et~al.
\newblock The llama 3 herd of models.
\newblock \emph{arXiv preprint arXiv:2407.21783}, 2024.

\bibitem[Fu et~al.(2023)Fu, Ou, Chen, Wan, Peng, and Khot]{fu2023chain}
Yao Fu, Litu Ou, Mingyu Chen, Yuhao Wan, Hao Peng, and Tushar Khot.
\newblock Chain-of-thought hub: A continuous effort to measure large language models' reasoning performance.
\newblock \emph{arXiv preprint arXiv:2305.17306}, 2023.

\bibitem[Fu et~al.(2024)Fu, Panda, Niu, Yue, Hajishirzi, Kim, and Peng]{fu2024data}
Yao Fu, Rameswar Panda, Xinyao Niu, Xiang Yue, Hannaneh Hajishirzi, Yoon Kim, and Hao Peng.
\newblock Data engineering for scaling language models to 128k context.
\newblock \emph{arXiv preprint arXiv:2402.10171}, 2024.

\bibitem[gkamradt(2023)]{NIAH}
gkamradt.
\newblock Llmtest-needleinahaystack.
\newblock \url{https://github.com/gkamradt/LLMTest_NeedleInAHaystack}, 2023.

\bibitem[Hendrycks et~al.(2020)Hendrycks, Burns, Basart, Zou, Mazeika, Song, and Steinhardt]{hendrycks2020measuring}
Dan Hendrycks, Collin Burns, Steven Basart, Andy Zou, Mantas Mazeika, Dawn Song, and Jacob Steinhardt.
\newblock Measuring massive multitask language understanding.
\newblock \emph{arXiv preprint arXiv:2009.03300}, 2020.

\bibitem[Hong et~al.(2024)Hong, Lee, and Thorne]{hong2024reference}
Jiwoo Hong, Noah Lee, and James Thorne.
\newblock Reference-free monolithic preference optimization with odds ratio.
\newblock \emph{arXiv preprint arXiv:2403.07691}, 2024.

\bibitem[Hsieh et~al.(2024)Hsieh, Sun, Kriman, Acharya, Rekesh, Jia, and Ginsburg]{hsieh2024ruler}
Cheng-Ping Hsieh, Simeng Sun, Samuel Kriman, Shantanu Acharya, Dima Rekesh, Fei Jia, and Boris Ginsburg.
\newblock Ruler: What's the real context size of your long-context language models?
\newblock \emph{arXiv preprint arXiv:2404.06654}, 2024.

\bibitem[Hu et~al.(2021)Hu, Shen, Wallis, Allen-Zhu, Li, Wang, Wang, and Chen]{hu2021lora}
Edward~J Hu, Yelong Shen, Phillip Wallis, Zeyuan Allen-Zhu, Yuanzhi Li, Shean Wang, Lu~Wang, and Weizhu Chen.
\newblock Lora: Low-rank adaptation of large language models.
\newblock \emph{arXiv preprint arXiv:2106.09685}, 2021.

\bibitem[Jiang et~al.(2023)Jiang, Sablayrolles, Mensch, Bamford, Chaplot, Casas, Bressand, Lengyel, Lample, Saulnier, et~al.]{jiang2023mistral}
Albert~Q Jiang, Alexandre Sablayrolles, Arthur Mensch, Chris Bamford, Devendra~Singh Chaplot, Diego de~las Casas, Florian Bressand, Gianna Lengyel, Guillaume Lample, Lucile Saulnier, et~al.
\newblock Mistral 7b.
\newblock \emph{arXiv preprint arXiv:2310.06825}, 2023.

\bibitem[Jin et~al.(2024)Jin, Han, Yang, Jiang, Liu, Chang, Chen, and Hu]{jin2024llm}
Hongye Jin, Xiaotian Han, Jingfeng Yang, Zhimeng Jiang, Zirui Liu, Chia-Yuan Chang, Huiyuan Chen, and Xia Hu.
\newblock Llm maybe longlm: Self-extend llm context window without tuning.
\newblock \emph{arXiv preprint arXiv:2401.01325}, 2024.

\bibitem[Lin et~al.(2021)Lin, Hilton, and Evans]{lin2021truthfulqa}
Stephanie Lin, Jacob Hilton, and Owain Evans.
\newblock Truthfulqa: Measuring how models mimic human falsehoods.
\newblock \emph{arXiv preprint arXiv:2109.07958}, 2021.

\bibitem[Meng et~al.(2024)Meng, Xia, and Chen]{meng2024simpo}
Yu~Meng, Mengzhou Xia, and Danqi Chen.
\newblock Simpo: Simple preference optimization with a reference-free reward.
\newblock \emph{arXiv preprint arXiv:2405.14734}, 2024.

\bibitem[Ouyang et~al.(2022)Ouyang, Wu, Jiang, Almeida, Wainwright, Mishkin, Zhang, Agarwal, Slama, Ray, et~al.]{ouyang2022training}
Long Ouyang, Jeffrey Wu, Xu~Jiang, Diogo Almeida, Carroll Wainwright, Pamela Mishkin, Chong Zhang, Sandhini Agarwal, Katarina Slama, Alex Ray, et~al.
\newblock Training language models to follow instructions with human feedback.
\newblock \emph{Advances in neural information processing systems}, 35:\penalty0 27730--27744, 2022.

\bibitem[Peng et~al.(2023)Peng, Quesnelle, Fan, and Shippole]{peng2023yarn}
Bowen Peng, Jeffrey Quesnelle, Honglu Fan, and Enrico Shippole.
\newblock Yarn: Efficient context window extension of large language models.
\newblock \emph{arXiv preprint arXiv:2309.00071}, 2023.

\bibitem[Press et~al.(2021)Press, Smith, and Lewis]{press2021train}
Ofir Press, Noah~A Smith, and Mike Lewis.
\newblock Train short, test long: Attention with linear biases enables input length extrapolation.
\newblock \emph{arXiv preprint arXiv:2108.12409}, 2021.

\bibitem[Rae et~al.(2019)Rae, Potapenko, Jayakumar, and Lillicrap]{rae2019compressive}
Jack~W Rae, Anna Potapenko, Siddhant~M Jayakumar, and Timothy~P Lillicrap.
\newblock Compressive transformers for long-range sequence modelling.
\newblock \emph{arXiv preprint arXiv:1911.05507}, 2019.

\bibitem[Rafailov et~al.(2024)Rafailov, Sharma, Mitchell, Manning, Ermon, and Finn]{rafailov2024direct}
Rafael Rafailov, Archit Sharma, Eric Mitchell, Christopher~D Manning, Stefano Ermon, and Chelsea Finn.
\newblock Direct preference optimization: Your language model is secretly a reward model.
\newblock \emph{Advances in Neural Information Processing Systems}, 36, 2024.

\bibitem[Raffel et~al.(2020)Raffel, Shazeer, Roberts, Lee, Narang, Matena, Zhou, Li, and Liu]{raffel2020exploring}
Colin Raffel, Noam Shazeer, Adam Roberts, Katherine Lee, Sharan Narang, Michael Matena, Yanqi Zhou, Wei Li, and Peter~J Liu.
\newblock Exploring the limits of transfer learning with a unified text-to-text transformer.
\newblock \emph{Journal of machine learning research}, 21\penalty0 (140):\penalty0 1--67, 2020.

\bibitem[Ravaut et~al.(2024)Ravaut, Sun, Chen, and Joty]{ravaut2024context}
Mathieu Ravaut, Aixin Sun, Nancy Chen, and Shafiq Joty.
\newblock On context utilization in summarization with large language models.
\newblock In \emph{Proceedings of the 62nd Annual Meeting of the Association for Computational Linguistics (Volume 1: Long Papers)}, pp.\  2764--2781, 2024.

\bibitem[Ru et~al.(2024)Ru, Qiu, Hu, Zhang, Shi, Chang, Cheng, Wang, Sun, Li, et~al.]{ru2024ragchecker}
Dongyu Ru, Lin Qiu, Xiangkun Hu, Tianhang Zhang, Peng Shi, Shuaichen Chang, Jiayang Cheng, Cunxiang Wang, Shichao Sun, Huanyu Li, et~al.
\newblock Ragchecker: A fine-grained framework for diagnosing retrieval-augmented generation.
\newblock \emph{arXiv preprint arXiv:2408.08067}, 2024.

\bibitem[Ruoss et~al.(2023)Ruoss, Del{\'e}tang, Genewein, Grau-Moya, Csord{\'a}s, Bennani, Legg, and Veness]{ruoss2023randomized}
Anian Ruoss, Gr{\'e}goire Del{\'e}tang, Tim Genewein, Jordi Grau-Moya, R{\'o}bert Csord{\'a}s, Mehdi Bennani, Shane Legg, and Joel Veness.
\newblock Randomized positional encodings boost length generalization of transformers.
\newblock \emph{arXiv preprint arXiv:2305.16843}, 2023.

\bibitem[Saeidi et~al.(2024)Saeidi, Verma, RRV, and Baral]{saeidi2024triple}
Amir Saeidi, Shivanshu Verma, Aswin RRV, and Chitta Baral.
\newblock Triple preference optimization: Achieving better alignment with less data in a single step optimization.
\newblock \emph{arXiv preprint arXiv:2405.16681}, 2024.

\bibitem[Schulman et~al.(2017)Schulman, Wolski, Dhariwal, Radford, and Klimov]{schulman2017proximal}
John Schulman, Filip Wolski, Prafulla Dhariwal, Alec Radford, and Oleg Klimov.
\newblock Proximal policy optimization algorithms.
\newblock \emph{arXiv preprint arXiv:1707.06347}, 2017.

\bibitem[Shi et~al.(2023)Shi, Chen, Misra, Scales, Dohan, Chi, Sch{\"a}rli, and Zhou]{shi2023large}
Freda Shi, Xinyun Chen, Kanishka Misra, Nathan Scales, David Dohan, Ed~H Chi, Nathanael Sch{\"a}rli, and Denny Zhou.
\newblock Large language models can be easily distracted by irrelevant context.
\newblock In \emph{International Conference on Machine Learning}, pp.\  31210--31227. PMLR, 2023.

\bibitem[Touvron et~al.(2023)Touvron, Martin, Stone, Albert, Almahairi, Babaei, Bashlykov, Batra, Bhargava, Bhosale, et~al.]{touvron2023llama}
Hugo Touvron, Louis Martin, Kevin Stone, Peter Albert, Amjad Almahairi, Yasmine Babaei, Nikolay Bashlykov, Soumya Batra, Prajjwal Bhargava, Shruti Bhosale, et~al.
\newblock Llama 2: Open foundation and fine-tuned chat models.
\newblock \emph{arXiv preprint arXiv:2307.09288}, 2023.

\bibitem[Tworkowski et~al.(2024)Tworkowski, Staniszewski, Pacek, Wu, Michalewski, and Mi{\l}o{\'s}]{tworkowski2024focused}
Szymon Tworkowski, Konrad Staniszewski, Miko{\l}aj Pacek, Yuhuai Wu, Henryk Michalewski, and Piotr Mi{\l}o{\'s}.
\newblock Focused transformer: Contrastive training for context scaling.
\newblock \emph{Advances in Neural Information Processing Systems}, 36, 2024.

\bibitem[Wu et~al.(2024{\natexlab{a}})Wu, Wang, Fu, Yue, Zhu, and Li]{wu2024long}
Wenhao Wu, Yizhong Wang, Yao Fu, Xiang Yue, Dawei Zhu, and Sujian Li.
\newblock Long context alignment with short instructions and synthesized positions.
\newblock \emph{arXiv preprint arXiv:2405.03939}, 2024{\natexlab{a}}.

\bibitem[Wu et~al.(2024{\natexlab{b}})Wu, Wang, Xiao, Peng, and Fu]{wu2024retrieval}
Wenhao Wu, Yizhong Wang, Guangxuan Xiao, Hao Peng, and Yao Fu.
\newblock Retrieval head mechanistically explains long-context factuality.
\newblock \emph{arXiv preprint arXiv:2404.15574}, 2024{\natexlab{b}}.

\bibitem[Xiong et~al.(2023)Xiong, Liu, Molybog, Zhang, Bhargava, Hou, Martin, Rungta, Sankararaman, Oguz, et~al.]{xiong2023effective}
Wenhan Xiong, Jingyu Liu, Igor Molybog, Hejia Zhang, Prajjwal Bhargava, Rui Hou, Louis Martin, Rashi Rungta, Karthik~Abinav Sankararaman, Barlas Oguz, et~al.
\newblock Effective long-context scaling of foundation models.
\newblock \emph{arXiv preprint arXiv:2309.16039}, 2023.

\bibitem[Xu et~al.(2024)Xu, Sharaf, Chen, Tan, Shen, Van~Durme, Murray, and Kim]{xu2024contrastive}
Haoran Xu, Amr Sharaf, Yunmo Chen, Weiting Tan, Lingfeng Shen, Benjamin Van~Durme, Kenton Murray, and Young~Jin Kim.
\newblock Contrastive preference optimization: Pushing the boundaries of llm performance in machine translation.
\newblock \emph{arXiv preprint arXiv:2401.08417}, 2024.

\bibitem[Yang et~al.(2024)Yang, Yang, Hui, Zheng, Yu, Zhou, Li, Li, Liu, Huang, Dong, Wei, Lin, Tang, Wang, Yang, Tu, Zhang, Ma, Xu, Zhou, Bai, He, Lin, Dang, Lu, Chen, Yang, Li, Xue, Ni, Zhang, Wang, Peng, Men, Gao, Lin, Wang, Bai, Tan, Zhu, Li, Liu, Ge, Deng, Zhou, Ren, Zhang, Wei, Ren, Fan, Yao, Zhang, Wan, Chu, Liu, Cui, Zhang, and Fan]{qwen2}
An~Yang, Baosong Yang, Binyuan Hui, Bo~Zheng, Bowen Yu, Chang Zhou, Chengpeng Li, Chengyuan Li, Dayiheng Liu, Fei Huang, Guanting Dong, Haoran Wei, Huan Lin, Jialong Tang, Jialin Wang, Jian Yang, Jianhong Tu, Jianwei Zhang, Jianxin Ma, Jin Xu, Jingren Zhou, Jinze Bai, Jinzheng He, Junyang Lin, Kai Dang, Keming Lu, Keqin Chen, Kexin Yang, Mei Li, Mingfeng Xue, Na~Ni, Pei Zhang, Peng Wang, Ru~Peng, Rui Men, Ruize Gao, Runji Lin, Shijie Wang, Shuai Bai, Sinan Tan, Tianhang Zhu, Tianhao Li, Tianyu Liu, Wenbin Ge, Xiaodong Deng, Xiaohuan Zhou, Xingzhang Ren, Xinyu Zhang, Xipin Wei, Xuancheng Ren, Yang Fan, Yang Yao, Yichang Zhang, Yu~Wan, Yunfei Chu, Yuqiong Liu, Zeyu Cui, Zhenru Zhang, and Zhihao Fan.
\newblock Qwen2 technical report.
\newblock \emph{arXiv preprint arXiv:2407.10671}, 2024.

\bibitem[Yang et~al.(2023)Yang, Wang, Shen, Panda, and Kim]{yang2023gated}
Songlin Yang, Bailin Wang, Yikang Shen, Rameswar Panda, and Yoon Kim.
\newblock Gated linear attention transformers with hardware-efficient training.
\newblock \emph{arXiv preprint arXiv:2312.06635}, 2023.

\bibitem[Yuan et~al.(2024)Yuan, Yuan, Tan, Wang, Huang, and Huang]{yuan2024rrhf}
Hongyi Yuan, Zheng Yuan, Chuanqi Tan, Wei Wang, Songfang Huang, and Fei Huang.
\newblock Rrhf: Rank responses to align language models with human feedback.
\newblock \emph{Advances in Neural Information Processing Systems}, 36, 2024.

\bibitem[Zhang(2024)]{zhang2024sinklora}
Hengyu Zhang.
\newblock Sinklora: Enhanced efficiency and chat capabilities for long-context large language models.
\newblock \emph{arXiv preprint arXiv:2406.05678}, 2024.

\bibitem[Zhang et~al.(2024{\natexlab{a}})Zhang, Bai, Lv, Gu, Liu, Zou, Cao, Hou, Dong, Feng, and Li]{zhang2024longcite}
Jiajie Zhang, Yushi Bai, Xin Lv, Wanjun Gu, Danqing Liu, Minhao Zou, Shulin Cao, Lei Hou, Yuxiao Dong, Ling Feng, and Juanzi Li.
\newblock Longcite: Enabling llms to generate fine-grained citations in long-context qa.
\newblock \emph{arXiv preprint arXiv:2409.02897}, 2024{\natexlab{a}}.

\bibitem[Zhang et~al.(2024{\natexlab{b}})Zhang, Shao, Liu, Xiao, Qian, Ye, and Dou]{zhang2024extending}
Peitian Zhang, Ninglu Shao, Zheng Liu, Shitao Xiao, Hongjin Qian, Qiwei Ye, and Zhicheng Dou.
\newblock Extending llama-3's context ten-fold overnight.
\newblock \emph{arXiv preprint arXiv:2404.19553}, 2024{\natexlab{b}}.

\bibitem[Zhu et~al.(2023)Zhu, Yang, Wang, Song, Wu, Wei, and Li]{zhu2023pose}
Dawei Zhu, Nan Yang, Liang Wang, Yifan Song, Wenhao Wu, Furu Wei, and Sujian Li.
\newblock Pose: Efficient context window extension of llms via positional skip-wise training.
\newblock \emph{arXiv preprint arXiv:2309.10400}, 2023.

\bibitem[Zhu et~al.(2024)Zhu, Guo, Shao, Yang, Wang, Xu, Wu, Li, Gao, Ma, et~al.]{zhu2024deepseek}
Qihao Zhu, Daya Guo, Zhihong Shao, Dejian Yang, Peiyi Wang, Runxin Xu, Y~Wu, Yukun Li, Huazuo Gao, Shirong Ma, et~al.
\newblock Deepseek-coder-v2: Breaking the barrier of closed-source models in code intelligence.
\newblock \emph{arXiv preprint arXiv:2406.11931}, 2024.

\end{thebibliography}
\bibliographystyle{iclr2025_conference}

\clearpage
\appendix
\clearpage
\section{Limitation and Future Work}
This paper presents an efficient preference optimization training strategy (LOGO) tailored for long-context alignment.
However, there are several limitations:
\begin{itemize}
    \item Due to resource constraints within the academic community, the evaluation of real-world testing sets in LongBench may be affected by the varying prompts selected by different studies, which can lead to significant discrepancies in results. Consequently, we are unable to directly replicate the results from other works
    \item As mentioned in the main body~(Sec.~\ref{subsec:logo_data}), there remains a lack of suitable evaluation models to assess whether the outputs of LCMs are accurate or contain hallucinations. The LOGO training objective proposed in this paper still has room for improvement.
    \item During the data construction phase, utilizing higher-quality datasets could yield better outcomes. However, as an academic paper, we believe we have demonstrated the generalizability of our method through the main experiments.
\end{itemize}

Moving forward, we plan to continue our research along the lines of efficient long-context alignment, particularly in algorithm development.
We aim to explore the integration of more effective evaluation strategies, such as RAG checkers~\citep{ru2024ragchecker}, to assist in constructing preference and dis-preference instances. Additionally, we should investigate how to enhance the efficiency of our LOGO data construction pipeline across various tasks and domains.

In summary, this paper highlights the substantial potential of efficient training in long-context scenarios, and we hope our work will provide valuable insights for future research endeavors.

\section{Details of Experiments in Introduction}
\label{appdix:into}
In this section, we introduce the preliminary studies in the Introduction section, including the experimental settings, task definitions, and retrieval score calculation. 

\paragraph{Experimental Settings}
In Fig.~\ref{fig:intro}(a) and Fig.~\ref{fig:intro}(b), we evaluate the model performance on the subsets in LongBench~\citep{bai2023longbench}, including Single Document QA, Multi-Document QA, Summarization, and Few-shot tasks.
For each long-context model, we utilize the same official instructions to guide the model prediction.

\paragraph{Multi-values Needle-in-a-Haystack} 
In Fig.~\ref{fig:intro}(c), we calculate the retrieval score on the Multi-values Needle-in-a-Haystack dataset, which requires LCMs to recall multiple values within the context. We provide an example in Fig.~\ref{fig:m_value_niah}:
\begin{figure}[h]
    \begin{AcademicBox}[\footnotesize Multi-values Needle-in-a-Haystack]
    \textbf{\textit{Context:}} \\
    \textcolor{gray!95}{... context ...} \\
    The best thing to do in San Francisco is \textcolor{red}{to eat a sandwich and sit in Dolores Park}.  \\
    \textcolor{gray!95}{... context ...} \\
    The best thing to do in New York is \textcolor{red}{to eat a sandwich and visit the Statue of Liberty.} \\
    \textcolor{gray!95}{... context ...} \\
    \vspace{-5pt} \hrule \vspace{4pt}
    \textbf{\textit{Question:}} \\
    What is the single best thing to do in both San Francisco and New York? \\
    \vspace{-5pt} \hrule \vspace{4pt}
    \textbf{\textit{Ground Truth: }} \textcolor{green}{(preference)} \\
    eat a sandwich 
    \end{AcademicBox}
    \vspace{-1em}
    \caption{Demonstration of Multi-values Needle-in-a-Haystack testing sample.}
    \label{fig:m_value_niah}
    \vspace{-1em}
\end{figure}

The formal definition of the task is as follows: Given $n$ questions $vq$ and its corresponding answers $K=\{vk_j\}_{j=1}^{n}$ (the needle),  we insert $K$ in a synthetic context $\vc$ (the haystack) at random position index ranges $P=\{vp_i\}_{i=1}^{n}$.
We then require the models to answer $\vq$ based on the haystack with the inserted needle.
It is worth noting that $\vq$ and $K$ are unique and irrelevant to the context,
ensuring that if an answer is correctly generated, it is indeed copied from the context, not from the model's internal knowledge.

\paragraph{Calculation of Retrieval Score}
Based on~\citet{wu2024retrieval}, we define the retrieval score as the recall score of salient tokens located by retrieval heads.
To enhance comprehension, we manage to utilize familiar symbols and definitions that align closely with previous research.
Specifically, denote the current token being generated during the auto-regressive decoding process as $x$, and the attention score of a head as $\boldsymbol{a} \in \mathcal{R}^{|\vc|}$.
In the task of Multi-values Needle-in-a-Haystack, an attention head $h$ is denoted as a retrieval head if it meets the following criteria:
\begin{itemize} \itemsep -0.1em
    \item $x \in \vk_i$, where $\vk_i \in K$ and $x$ is a token within any one of the needle sentences in $K$.
    \item $\vc_j= x$, $j = \arg\max(\boldsymbol{a})$, $j \in \vp_i$, $\vp_i \in P$, i.e., the input token that receives the highest attention probability by this head is a token within any one of the needle in $K$ and is the same token as the currently generated token. 
\end{itemize}

Let $\vg_h$ be the set containing all copy tokens~(also can be viewed as the located tokens) and pasted by a given head $h$, we define:
\begin{equation}
    \text{Retrieval score for head}\;\; h = \frac{|\vg_h \cap \vk_i |}{|\vk_i|}, 
\end{equation}
It is worth noting that the retrieval score represents a token-level recall rate of the most attended tokens by an attention head.
After obtaining the retrieval score for each head, we start by filtering out the non-retrieval heads by setting the threshold at 0.1. This means that if a head performs copy-paste 10\% of the time or more, it will be considered a retrieval head. 
Then, we calculate the retrieval head score by averaging the scores of the top 10 attention heads from the remaining retrieval heads.

% \textbf{Retrieval Head Detection Algorithm}\quad\quad
% We calculate the retrieval score for all attention heads under a diverse set of input contexts. 
% For each language model we consider, we compile three sets of Needle-in-a-Haystack samples, each consisting of a unique tuple $(\vq, \vk, \vx)$.
% For each sample, we make sure $(\vq, \vk)$ is semantically irrelevant with $\vx$ and that $\vq$ cannot be answered using the model's existing knowledge by manually inspecting the model output. 
% Then for each $(\vq, \vk, \vx)$ sample, we perform  Needle-in-a-Haystack on 20 different length values uniformly sampled from 1K-50K, where in each length, $\vq$ is inserted in 10 different depth uniformly ranging from the start to the end of $\vx$.
% We note that this scale of tests gives stable outputs as the average retrieval score converges after just a few samples. 
% In total, each language model is subjected to approximately 600 instances of retrieval testing. 
% We calculate the retrieval score for each attention head in each test and use the average of these scores as the head's final retrieval score.
% The attention heads with relatively larger retrieve score can be considered as retrieval heads.
% In our case (Fig.~\ref{fig:score_range}), we set the threshold as 0.1, meaning that as long as the head performs copy-paste 10\% of the times, we consider it a retrieval head.

\section{Design of LOGO Training Objective and Error Pattern Definition in LCMs}
\label{appdix:design_logo_train_obj}
\begin{figure}[h]
    \centering
    \includegraphics[width=1\linewidth]{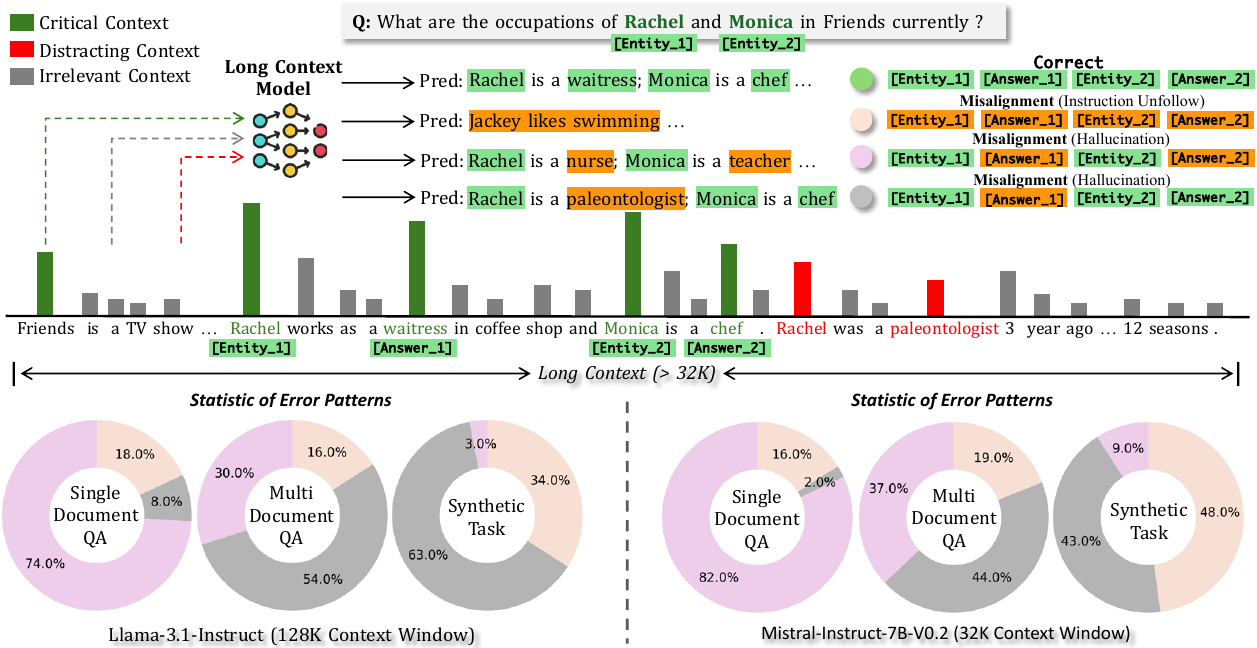}
    \caption{Demonstration and statistical analysis of different error patterns in long context tasks, where we have the following definitions of misalignment: (1) Instruction Unfollow: The entities in the model's prediction are different from the entities in the question; (2)  Hallucination: The entities in the prediction overlaps with the entities in the question, but the answer is incorrect.}
    \label{fig:define_error_pattern}
\end{figure}

Misaligned predictions generated from LCMs can be specifically categorized into two types: failing to follow instructions and generating hallucinations. 
In Fig.~\ref{fig:define_error_pattern}, we illustrate these two error patterns. Specifically, we define different error patterns by utilizing the degree of overlap between entities in the responses and the questions, along with specific templates:
\begin{itemize} \itemsep -0.1em
    \item \textbf{Instruction Unfollow}: the entities in the model's responses do not overlap with the entities in the question.
    \item \textbf{Hallucination}: there is a partial intersection of entities between the model’s responses and the question, and the entities in the response coincide with the main subject of the question.
\end{itemize}

It is worth mentioning that merely utilizing Named Entity Recognition~(NER) models and rule-based methods proves inadequate for identifying these patterns.
Instead, a more robust evaluation involving strong LLMs such as GPT-4 or human assessment is required to accurately identify these patterns.
Consequently, in the design of the LOGO training objective, we do not confine to constructing cases with specific error patterns. 
Therefore, instead of finding one dis-preference instance with a specific error pattern, we can expand the dis-preference space to push the model away from a range of possible dis-preference instances.

\label{appdix:error_pattern}

\section{Positional Indices Synthesis Details}
\label{appdix:pos_synthesis}

\begin{figure}[h]
\centering    
\subfigure[Continuous Chunk Positional Indices Synthesis]{
    \label{fig:pos_syn_a}     
    \includegraphics[width=0.9\linewidth]{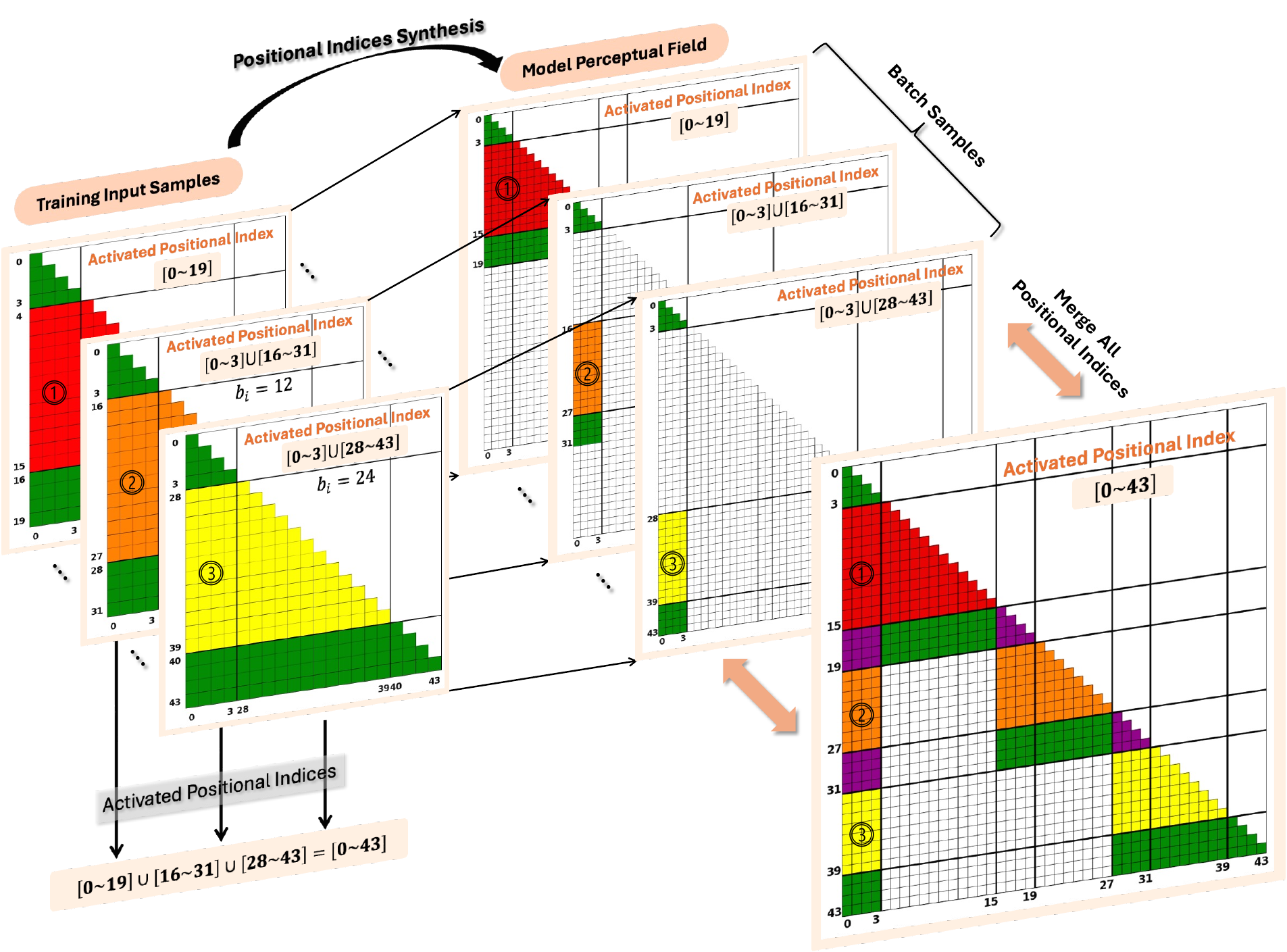}  
}     
\subfigure[Sparse Chunk Positional Indices Synthesis]{ 
    \label{fig:pos_syn_b}     
    \includegraphics[width=0.9\linewidth]{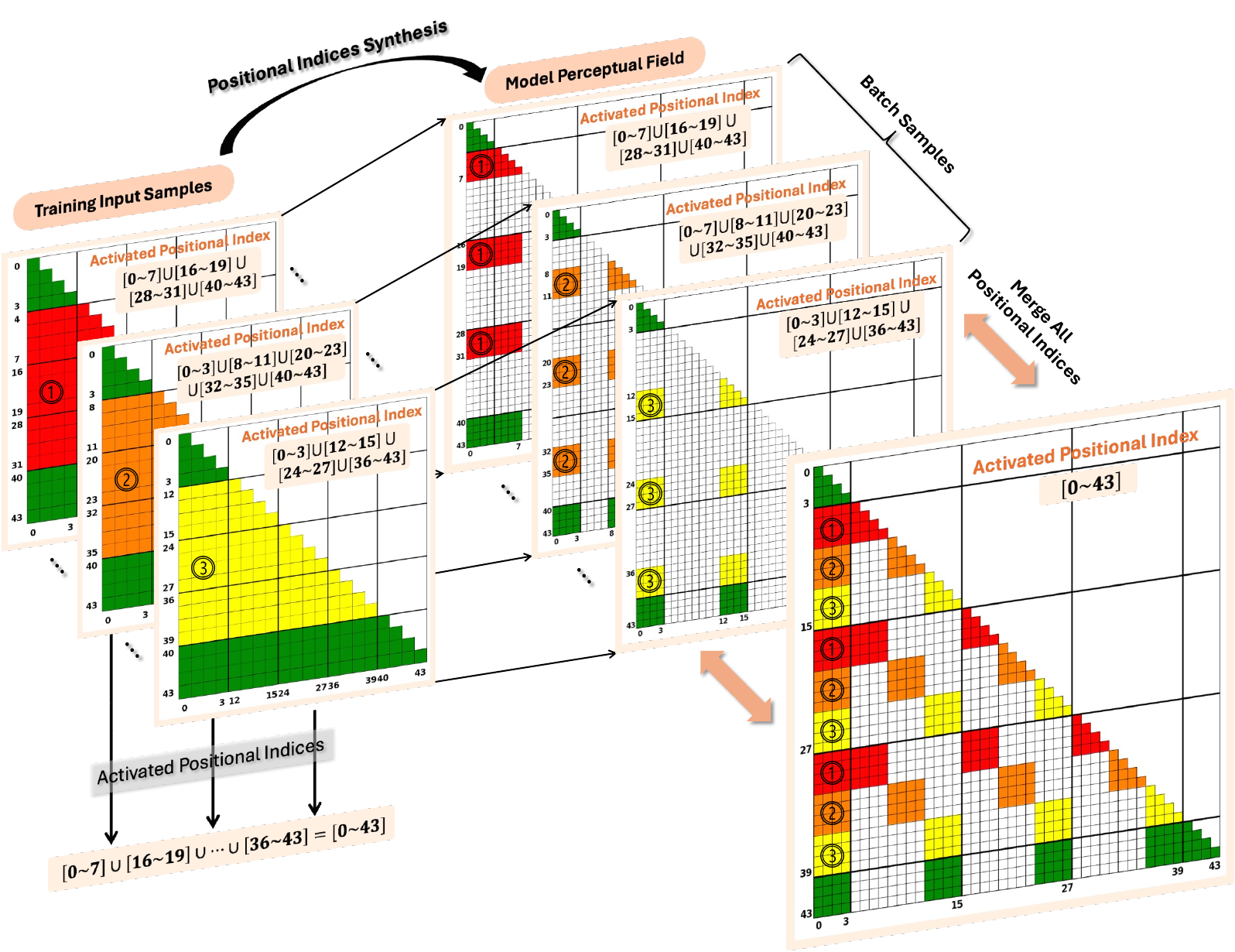}     
}    
\caption{Illustration of positional indices synthesis process, where the original context length is 19, and we extend it to a context length of 43. It is noteworthy that one batch in the figure corresponds to one training sample of LOGO, which contains one preference instance and several dis-preference instances.}     
\label{fig:pos_synthesis}     
\end{figure}

We visualize the positional indices synthesis process in Fig.~\ref{fig:pos_synthesis}.
Specifically, to ensure that the synthesized positional indices do not disrupt the original text's semantic structure while maximizing the extended context size, we employ two different strategies for positional bias $\mathcal{B}$: Continuous Chunk Positional Indices Synthesis~(Fig.~\ref{fig:pos_syn_a}) and Sparse Chunk Positional Indices Synthesis~(Fig.~\ref{fig:pos_syn_b}).
For Continuous Chunk Positional Indices Synthesis, the positional bias within the same chunk is consistent. For instance, in the first chunk $C_{0}$, the positional bias $\{b_0, b_1, \cdots, b_{|C_{i}|}\}$ are the same value sampled from distribution $\mathcal{U}(1, (K-k)/N)$. 
This ensures that the semantic structure within the chunk remains intact but can lead to sparse synthesized positional indices, as there will be significant gaps between the positional indices among different chunks. 
Thereby, we propose Sparse Chunk Positional Indices Synthesis to fill these gaps, where each positional bias $b_i$ is sampled uniformly according to Equ.~\ref{equ:pos_syn}. 
Considering that Sparse Chunk Positional Indices Synthesis might disrupt the semantic structure of the text, we set the ratio of data for Continuous Chunk Positional Indices Synthesis and Sparse Chunk Positional Indices Synthesis to 9:1 in actual deployment.

\section{Case Study of LOGO Data}
\label{appdix:case_study}
In this section, we provide the training samples built based on the LOGO training data construction pipeline as illustrated in Sec.~\ref{subsec:logo_data}.
We present the training samples in Fig.~\ref{fig:case1}, Fig.~\ref{fig:case2}, Fig.~\ref{fig:case3}, 
and Fig.~\ref{fig:case4}, where the training data exhibits different error patterns~(misalignments) in their dis-preference instances. 

\begin{figure}[h]
    \begin{AcademicBox}[\footnotesize Case Study 1]
    \textbf{\textit{Context:}} \\
    \textcolor{gray!95}{... context ...} \\
    If class time is used to allow students to apply what they have learned in cooperation with their peers, then \textcolor{red}{the Flipped Classroom Model encourages collaboration, inquiry, and problem solving}. \\
    \textcolor{gray!95}{... context ...} \\
    \vspace{-5pt} \hrule \vspace{4pt}
    \textbf{\textit{Question:}} \\
    Why is it important for teachers to use student-centered activities in the flipped classroom model? \\
    \vspace{-5pt} \hrule \vspace{4pt}
    \textbf{\textit{Ground Truth: }} \textcolor{green}{(preference)} \\
    To encourage communication, inquiry, research, problem solving, and collaboration. \\
    \vspace{-5pt} \hrule \vspace{4pt}
    \textbf{\textit{Dis-preference Prediction1: } \textcolor{red}{(Instruction Unfollowing)}} \\
    How can teachers create student-centered learning experiences in the classroom? What are some possible challenges in creating student-centered learning experiences in the classroom?  \\
    \vspace{-5pt} \hrule \vspace{4pt}
    \textbf{\textit{Dis-preference Prediction2: } \textcolor{red}{(‌Hallucination)}} \\
    Blended learning in action: A practical guide toward sustainable change. Corwin. Please note that you need to have at least a basic subscription to access the full text of the references... 
    \end{AcademicBox}
    \vspace{-1em}
    \caption{LOGO training samples: case1.}
    \label{fig:case1}
    \vspace{-1em}
\end{figure}

\begin{figure}[h]
    \begin{AcademicBox}[\footnotesize Case Study 2]
    \textbf{\textit{Context:}} \\
    \textcolor{gray!95}{... context ...} \\
    As always in quantum physics, the joint measurement on Alice's \textcolor{red}{two photons, concerning their relative state, produces a truly random result}, just one among several possibilities.   \\
    \textcolor{gray!95}{... context ...} \\
    Alice's joint measurement thus produces a random result. Depending on this result, Bob's photon will always produce the result that the initial photon would have produced if it had been measured in the same direction, or else it will produce the opposite result to the one that would have been produced by the initial photon. \\
    \textcolor{gray!95}{... context ...} \\
    \vspace{-5pt} \hrule \vspace{4pt}
    \textbf{\textit{Question:}} \\
    What is the significance of the energy correlation between the two infrared photons? \\
    \vspace{-5pt} \hrule \vspace{4pt}
    \textbf{\textit{Ground Truth: }} \textcolor{green}{(preference)} \\
    If one photon has an energy above the average, the other photon will necessarily have an energy below the average. \\
    \vspace{-5pt} \hrule \vspace{4pt}
    \textbf{\textit{Dis-preference Prediction1: } \textcolor{red}{(Instruction Unfollowing)}} \\
    Bell's game is a theoretical game where two players, Alice and Bob, make independent choices of the polarization state of two photons before measuring the correlations between their polarizations... \\
    \vspace{-5pt} \hrule \vspace{4pt}
    \textbf{\textit{Dis-preference Prediction2: } \textcolor{red}{(Instruction Unfollowing)}} \\
    None, since there is no mention of energy correlation between two infrared photons in the provided context. The context only discusses quantum nonlocality, Bell\'s game, and quantum teleportation.... 
    \end{AcademicBox}
    \vspace{-1em}
    \caption{LOGO training samples: case2.}
    \label{fig:case2}
    \vspace{-1em}
\end{figure}

\begin{figure}[h]
    \begin{AcademicBox}[\footnotesize Case Study 3]

    \textbf{\textit{Context:}} \\
    \textcolor{gray!95}{... context ...} \\
    What is the slope of the line containing points (2, –3) and (4, 3)? slope = 3  \\
    \textcolor{gray!95}{... context ...} \\
    A slope that goes from low to high is positive. A slope that goes from high to low is negative. A slope that goes straight across is 0. \\
    \textcolor{gray!95}{... context ...} \\
    \vspace{-5pt} \hrule \vspace{4pt}
    \textbf{\textit{Question:}} \\
    What is the slope of the line containing points (2, –3) and (4, 3)? \\
    \vspace{-5pt} \hrule \vspace{4pt}
    \textbf{\textit{Ground Truth: }} \textcolor{green}{(preference)} \\
    3 \\
    \vspace{-5pt} \hrule \vspace{4pt}
    \textbf{\textit{Dis-preference Prediction1: } \textcolor{red}{(Instruction Unfollowing)}} \\
    Do you want to solve it step by step? \\
    \vspace{-5pt} \hrule \vspace{4pt}
    \textbf{\textit{Dis-preference Prediction2: } \textcolor{red}{(Hallucination)}} \\
    The question can be interpreted as the difference in y-values divided by the difference in x-values. That is, let y1 = –3 and y2 = 3. Then x1 = 2 and x2 = 4. The slope is: = 6 
    \end{AcademicBox}
    \vspace{-1em}
    \caption{LOGO training samples: case3.}
    \label{fig:case3}
    \vspace{-1em}
\end{figure}

\begin{figure}[h]
    \begin{AcademicBox}[\footnotesize Case Study 4]
    \textbf{\textit{Context:}} \\
    \textcolor{gray!95}{... context ...} \\
    And then there was \textcolor{red}{no more speaking of the old lady}, the house with the candle and the woman who was like a candle until, in the last train, they were nearly home.  \\
    \textcolor{gray!95}{... context ...} \\
    \vspace{-5pt} \hrule \vspace{4pt}
    \textbf{\textit{Question:}} \\
    Why does the narrator stop asking his father about the grapes and the grape-vine? \\
    \vspace{-5pt} \hrule \vspace{4pt}
    \textbf{\textit{Ground Truth: }} \textcolor{green}{(preference)} \\
    Because of the unpleasant encounter with the old lady at the house. \\
    \vspace{-5pt} \hrule \vspace{4pt}
    \textbf{\textit{Dis-preference Prediction1: } \textcolor{red}{(Hallucination)}} \\
    The narrator stops asking his father about the grapes and the grape-vine because his father is not responding to the questions. The narrator is trying to learn about the grape-vine and its whereabouts but his father is not willing to provide any information...  \\
    \vspace{-5pt} \hrule \vspace{4pt}
    \textbf{\textit{Dis-preference Prediction2: } \textcolor{red}{(Hallucination)}} \\
    Blended learning in action: Because his father\'s eyes seemed to fill with water and he began to speak in a small voice. At the same time, the narrator himself felt like his heart had been "torn in half."... 
    \end{AcademicBox}
    \vspace{-1em}
    \caption{LOGO training samples: case4.}
    \label{fig:case4}
    \vspace{-1em}
\end{figure}

% \begin{figure}[h]
%     \begin{AcademicBox}[\footnotesize In-Context Templates for Toxic Detection]
%     \textbf{\textit{Demonstration 1}} \\
%     \# Input: None of this would be possible if we didn't have a fucked up \\
%     \# Output: fucked up \\
%     \textbf{\textit{Demonstration 2}} \\
%     \# Input: Thus when shit hits the fan insects will prove to be a \\
%     \# Output: shit \\
%     \textbf{\textit{Demonstration 3}} \\
%     \# Input: It should be noted, since idiots always misunderstand this point, that freedom \\
%     \# Output: idiots \\
%     \textbf{\textit{Demonstration 4}} \\
%     \# Input: For five years she was on and off with a sexist pig who would make her \\
%     \# Output: sexist pig
%     \end{AcademicBox}
%     \vspace{-1em}
%     \caption{In-context templates for toxic detection.}
%     \label{fig:in-context-learning-detect}
%     \vspace{-1em}
% \end{figure}

% \paragraph{Hyper-parameters for Positional Indices Synthesis}

% \section{Training Infrastructure and Details}
% \label{appdix:training_details}

\end{document}